\documentclass[twocolumn,letter, 8p]{autart}    
\usepackage{graphicx}
\usepackage{subfigure}
\usepackage[english]{babel}
\usepackage{amsmath}
\usepackage{color}
\usepackage{cleveref}
\usepackage{dsfont}
\usepackage{enumerate}
\usepackage{url}
\usepackage{cases}
\usepackage{arydshln}    

\crefname{figure}{Figure}{Figures}
\crefname{section}{Section}{Sections}
\crefname{table}{Table}{Tables}
\crefname{example}{Example}{Examples}
\crefname{definition}{Definition}{Definitions}

\newcommand{\bma}{\left[ \begin{array}}
\newcommand{\ema}{\end{array} \right]}
\newtheorem{theorem}{Theorem}

\newtheorem{definition}{Definition}
\newtheorem{lemma}{Lemma}

\newtheorem{assumption}{Assumption}
\crefname{proposition}{Proposition}{Propositions}
\crefname{theorem}{Theorem}{Theorems}
\crefname{lemma}{Lemma}{Lemmas}
\crefname{remark}{Remark}{Remarks}
\crefname{appendix}{Appendix}{Appendices}

\makeatletter \let\cl@part\relax \makeatother  

\begin{document}

\begin{frontmatter}

\title{Switching control for tracking of a hybrid position-force trajectory}

\thanks[footnoteinfo]{This research is supported by the Dutch Technology Foundation (STW). This paper was not presented at any IFAC meeting. Corresponding author D.J.F.~Heck}

\author{D.J.F.~Heck}\ead{d.j.f.heck@tue.nl},      
\author{A.~Saccon,}
\author{N.~van~de~Wouw,}
\author{H.~Nijmeijer}

\address{Eindhoven University of Technology, Department of Mechanical Engineering, P.O. Box 513, NL 5600 MB Eindhoven, The Netherlands}  


\begin{abstract}                          
This work proposes a control law for a manipulator with the aim of realizing desired time-varying motion-force profiles in the presence of a stiff environment.
In many cases, the interaction with the environment affects only one degree of freedom of the end-effector of the manipulator. Therefore, the focus is on this contact degree of freedom, and a switching position-force controller is proposed to perform the hybrid position-force tracking task.
Sufficient conditions are presented to guarantee input-to-state stability of the switching closed-loop system with respect to perturbations related to the time-varying desired motion-force profile.
The switching occurs when the manipulator makes or breaks contact with the environment.
The analysis  shows that to guarantee closed-loop stability while tracking arbitrary time-varying motion-force profiles, the controller should implement a considerable (and often unrealistic) amount of damping, resulting in inferior tracking performance.
Therefore, we propose to redesign the manipulator with a compliant wrist.
Guidelines are provided for the design of the compliant wrist while employing the designed switching control strategy, such that stable tracking of a motion-force reference trajectory can be achieved and bouncing of the manipulator while making contact with the stiff environment can be avoided. Finally, numerical simulations are presented to illustrate the effectiveness of the approach.

\end{abstract}

\end{frontmatter}

\section{Introduction} \label{sec: introduction}
Numerous applications such as, e.g., bilateral teleoperation, automated assembly tasks, and surface finishing involve the interaction between a robot manipulator and a stiff environment.
In those applications, the stability of transitions from free motion to constrained motion and from  constrained motion to free motion is essential for accomplishing the desired task. Ensuring stability during these transitions is a challenge as the combined robot-environment dynamics switches abruptly at the moment of contact and detachment from the environment.

Different control architectures have been proposed for motion-force control of a manipulator in contact (for an overview, see, e.g., \cite[Chapter 7]{SHoR2008}), but the stability question is still open.
The most studied and applied control schemes include stiffness, impedance and admittance control \cite{Hogan1988,Volpe1993,CanudasdeWit1997,Jung2004,ZovoticStanisic2012,Ge2014},
hybrid position-force control \cite{Raibert1981,Khatib1987},
and parallel position-force control \cite{Chiaverini1993}.
The gains in these control schemes are tuned separately for free motion and constrained motion. Stability of the resulting closed-loop dynamics is analyzed using standard Lyapunov methods and guaranteed for free motion and constrained motion, but the contact and detachment transitions are not included in the analysis. Bouncing and unstable contact behavior might therefore still occur, and do still occur. As a practical solution, when implementing these control schemes on a physical manipulator, the manipulator is usually commanded to approach the environment with a very slow velocity to prevent the excitation of the unstable contact dynamics.

The aim of this paper is to go beyond the current state of the art and propose a novel stability analysis for this problem. We propose a mathematical analysis that can help
control engineers as well as mechanical designers to develop controlled manipulators that
exhibit stable contact behavior with a stiff environment. We propose a controller and a stability analysis to verify if stability is guaranteed even during contact and detachment phases.
A key aspect is that we are interested in tracking {\em time-varying} motion and force profiles. This specific goal originates from our interest in telerobotics, where a force and position reference from the master device has to be translated into a command for the slave device. As the force and position reference comes from a human operator, we want to allow those reference signals to be as general as possible.

Few theoretical studies have addressed directly the root cause of the instability during bouncing against a stiff environment. In \cite{Tarn1996,Doulgeri2005}, a switched position-force controller is considered, where the controller switches from motion to force control when contact with the environment is made. Using analysis techniques for switched systems, conditions for asymptotic stability are derived for a \emph{constant} position or force setpoint regulation problem.
Hysteresis switching is considered in \cite{Carloni2007} to prevent bouncing of the manipulator against the environment. In \cite{ZovoticStanisic2012}, a switching rule is designed for the impedance parameters to dissipate the kinetic energy engaged at impact.
The resulting ``active impedance control'' guarantees ``velocity regulation in free motion, impact attenuation'' and tracking of a \emph{constant} force setpoint in contact.
The number of bounces is cleverly minimized in \cite{Pagilla2001} by exploiting a transition controller, but then the contact force is controlled to a \emph{constant} setpoint.
In \cite{Lai2012}, nonlinear damping is proposed to minimize the force overshoot without compromising the settling time. In all these publications, tracking of desired \emph{time-varying} motion and force profiles is not considered.

In the above mentioned papers, the manipulator-environment interaction is modeled using a flexible spring-damper contact model. The stiffness and damping properties of the
environment are included explicitly and, as a consequence, the impact phase has a finite time duration.
Such a modeling approach is also taken in this paper.

Manipulator-environment interaction can also be modeled using tools from nonsmooth mechanics \cite{Brogliato1999,Leine2008}. In doing so, the time duration of the impact event is assumed to be zero and an impact law (e.g., Newton's law of restitution) is employed to characterize the collision. Stable tracking of specific force/position profiles using such nonsmooth mechanics modeling formalism has been addressed in this context. 
In \cite{Pagilla2001a}, a discontinuous control scheme is proposed to ensure stable regulation on the surface of the unilateral constraint. A switched motion-force tracking controller for manipulators subject to unilateral constraints is considered in \cite{Brogliato1997,Bourgeot2005,Morarescu2010}. There, it is shown that the design of the desired trajectory in the transition phase is crucial for achieving stability.

To the best of authors' knowledge, the problem of stable tracking of arbitrary force/position profiles as we consider in this work has not been solved even in the framework of nonsmooth mechanics. The stability of the tracking controller cast in this framework is clearly of interest and deserves further investigation.
This framework will not be addressed here just because, as we mentioned, we adopt a flexible (spring-damper) contact model. 


In this work, we propose a control law for making a manipulator track  a \emph{time-varying} motion and force profile.
Because in many tasks of practical interest the interaction of the robot end-effector with the environment occurs just in one direction, we study the contact stability problem using a 1-DOF dynamic manipulator model.
The remaining unconstrained DOFs can be controlled with standard motion control techniques (see \cite{Spong2006}).
We propose a switched motion-force tracking control strategy and include the transitions from free motion to contact (and vice versa) in the stability analysis of the closed-loop dynamics.
The obtained stability conditions are given in \cref{proposition: GUES} in \cref{sec: stability analysis}.
The stability analysis of the closed-loop system reveals that the controller should implement a considerable amount of damping to guarantee stability while tracking an arbitrary time-varying motion-force profile.
Because an excessive amount of damping limits the tracking performance due to a sluggish response,
we propose an alternative mechanical manipulator design by including a compliant wrist.
In this way, the resonance frequency of the impact and contact transients can be reduced and the associated energy can be dissipated in a passive way.
The use of such an ``energy absorbing component'' is mentioned in \cite{Volpe1993}, but a stability analysis is not considered therein.

The main contributions of this paper are as follows.
First, we propose a
combination of the compliant wrist design with a novel switched motion-force controller for the tracking of time-varying motion and force profiles.
Secondly, we propose a stability analysis that provides design guidelines for both the compliant wrist and controller to guarantee stable contact while tracking arbitrary motion and force profiles.
In particular, we show how bouncing of the manipulator against the stiff environment can be prevented without the need of a considerable amount of damping from the controller.

This article is organized as follows. In \cref{sec: hybrid system}, the manipulator and environment model are introduced and the  controller design is proposed. The stability analysis is described in \cref{sec: stability analysis}. \cref{sec: stiff environment} illustrates the obtained results by means of a simulation study. \cref{sec: compliant manipulator} discusses the benefits of additional (wrist-)compliance in the manipulator and the conclusions are presented in \cref{sec: conclusion}.

\section{System modeling and controller design} \label{sec: hybrid system}
Our primary goal is to design a controller for making a manipulator track a desired motion-force profile.
As explained in the introduction, we focus on a 1-DOF modeling of the manipulator-environment interaction.

\begin{figure}
    \centering
    \includegraphics[width = 4.5cm]{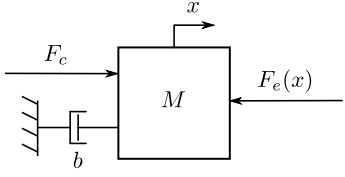}
    \caption{1-DOF manipulator.}
    \label{fig: block diagram manipulator}
\end{figure}

Consider the decoupled contact DOF of the manipulator  as depicted in \cref{fig: block diagram manipulator}. The Cartesian space dynamics are described by
\begin{equation}
  M \ddot{x} + b \dot{x}= F_c - F_e, \label{dynamics manipulator}
\end{equation}
where $x$ represents the manipulator position,
$M>0$ the equivalent mass of the manipulator,
$b>0$ the viscous friction in the joint,
$F_c$ the control force and
$F_e$ the force exchanged between the environment and the manipulator.
The environment is modeled as a static wall  at $x=0$ and, without loss of generality,  the manipulator is in contact with the environment for $x>0$.
In \cite{Tarn1996,Doulgeri2005}, the environment is modeled as a piecewise linear spring.
We consider, similarly to \cite{Carloni2007}, an extended model including damping and friction. Namely, we use the Kelvin-Voigt contact model
\begin{equation}
  F_e(x,\dot x) = \left\{ \begin{array}{ll}
     0 & \mbox{ for } x \leq 0 \\
     k_e x + b_e \dot{x} & \mbox{ for } x > 0
  \end{array} \right.  \label{Kelvin-Voigt}
\end{equation}
with $k_e>0$ and $b_e>0$ the stiffness and damping properties of the environment, respectively.
This model is nonlinear and non-smooth due to the abrupt change in $F_e$ at $x=0$.

In free motion, the manipulator is required to follow a bounded desired motion profile $x_d(t)$, whereas in contact, a desired force profile $F_d(t)$ should be applied to the environment.
Impedance controllers have been proposed in, e.g., \cite{CanudasdeWit1997,Jung2004} to control the contact force $F_e$ by creating a force loop around an inner motion control loop.
In this way, a desired impedance of the contact is designed, but the contact force is controlled indirectly.
We propose, instead, the following switched motion-force controller that switches between a resolved acceleration controller in free motion and a
force controller in the contact phase:

\vspace{-3.5mm}
\small
\begin{subequations}
\label{Fc}
\begin{numcases}{\hspace{-4mm}F_c\hspace{-1mm}=\hspace{-2mm}}
  \hspace{-2mm}M_a \ddot{x}_d(t) + k_d (\dot{x}_d(t) - \dot{x}) + k_p (x_d(t) - x ), &\hspace{-3.5mm}$\forall x \leq 0$,\label{controlller free motion} \vspace{2mm}\\
  \hspace{-2mm}F_d(t) + k_f (F_d(t)-F_e) - b_f \dot{x}, &\hspace{-3.5mm}$\forall x > 0 $, \label{controller contact}
\end{numcases}
\end{subequations}
\normalsize
such that both motion and force are controlled directly.
Here, $k_p>0$ and $k_d>0$ are the proportional and derivative gains of the motion controller, respectively. The estimated mass of the manipulator $M_a>0$ in \eqref{controlller free motion} might differ from the actual mass $M$ in \eqref{dynamics manipulator} due to uncertainties in the model parameter identification. The gain $k_f>0$ represents the proportional term of the force controller and $b_f>0$ is the damping gain, dissipating energy during the contact phase.
For the controller \eqref{Fc}, it is assumed that the contact force $F_e$, position $x$ and velocity $\dot{x}$ can be measured.
Although, in \eqref{Fc}, the switching between motion control and force control is decided based on the actual position $x$ of the manipulator, for a stiff environment, $k_e \gg b_e$, this is equivalent to switching based on the interaction force $F_e$. This implies that a perfect knowledge of the location of the environment is not necessary for the implementation of the controller defined by \eqref{Fc}.

In order to analyze stability of the system described by \eqref{dynamics manipulator}-\eqref{Fc}, we reformulate the closed-loop dynamics as a switching state-space model.
A key idea for the stability analysis, detailed in \cref{sec: stability analysis}, is to express the force tracking error $F_d(t)-F_e$ in terms of the motion tracking error $x_d(t)-x$,
such that both in free motion and in contact the goal is to make the tracking error $x_d(t)-x$ small.
In contact, $x_d(t)$ then represents the 'virtual' desired trajectory, corresponding to the desired contact force $F_d(t)$.
For the relationship between $F_d(t)$ and $x_d(t)$ during contact,
$x \rightarrow x_d(t)$ should also imply $F_e \rightarrow F_d(t)$. To this end, we consider the following relationship to deduce $x_d(t)$ from $F_d(t)$ in the contact phase:
\begin{equation}
 \hat{k}_e x_d(t) + \hat{b}_e \dot{x}_d(t) = F_d(t), \quad \mbox{for } F_d(t)>0,  \label{Fd hat}
\end{equation}
where $\hat{k}_e$ and $\hat{b}_e$ are available estimates of $k_e$ and $b_e$. 
\begin{assumption} \label{assumption: xd}
 The desired position $x_d(t)$ and velocity $\dot{x}_d(t)$ trajectories are continuous, and the desired acceleration $\ddot{x}_d(t)$ is piecewise-continuous and bounded.
\end{assumption}

Two separate user-defined motion and force profiles can be glued together to satisfy \cref{assumption: xd} by using the design procedure detailed in Appendix \ref{appendix: design desired trajectories}.

In terms of the exact parameters $k_e$ and $b_e$, \eqref{Fd hat} can be rewritten as
\begin{equation}
  k_e x_d(t) + b_e \dot{x}_d(t) + w_f(t) = F_d(t), \quad \mbox{for } F_d(t)>0, \label{Fd}
\end{equation}
with $w_f(t):=(\hat{k}_e-k_e) x_d(t) + (\hat{b}_e-b_e) \dot{x}_d(t)$ a bounded --due to \cref{assumption: xd}-- perturbation.
When the estimates $\hat{k}_e$ and $\hat{b}_e$ are exact, $w_f(t) = 0$ and $x - x_d(t) \rightarrow 0$ implies that $F_e - F_d(t) \rightarrow 0$. When $\hat{k}_e \neq k_e$ and/or $\hat{b}_e \neq b_e$, $w_f(t) \neq 0$ and acts a perturbation in the stability analysis. Since the mapping \eqref{Fd} is only used for the stability analysis and not in the controller \eqref{Fc}, the lack of exact knowledge of $k_e$ and $b_e$ will not affect the stability or tracking of the system described by \eqref{dynamics manipulator}-\eqref{Fc}.

The tracking error
\begin{equation}
  z = \bma{c} z_1 \\ z_2 \ema := \bma{c} x_d(t)-x \\ \dot{x}_d(t) - \dot{x} \ema \label{states}
\end{equation}
can be used to rewrite the closed-loop system dynamics \eqref{dynamics manipulator}-\eqref{Fc} and \eqref{Fd} as the following perturbed switched system
\begin{eqnarray}
   & \Sigma^p: \quad
  \dot{z} = A_i z + N w_i(t) = \bma{cc} 0 & 1 \\ -K_i & -B_i \ema z + N w_i(t), &\nonumber \\
   & z\in \Omega_i(t), i \in \{1,2\}, &\label{switched system general structure}
\end{eqnarray}
where $N = \left[0,~1 \right]^T$ and
\begin{subequations}
 \label{K1 K2 B1 B2 w1 w2}
 \begin{eqnarray}
 &\hspace{-0mm}&K_1:=\frac{k_p}{M}, \quad  B_1:= \frac{k_d+b}{M},  \label{K1 and B1}\\
 \hspace{-7mm}
 &\hspace{-0mm}&K_2:= \frac{(1+k_f)k_e}{M}, \quad B_2:= \frac{(1+k_f)b_e+b_f+b}{M}, \label{K2 and B2} 
 \end{eqnarray}
 \begin{eqnarray}
  &\hspace{-0mm}& w_1(t) := \frac{M-M_a}{M} \ddot{x}_d(t) + \frac{b}{M} \dot{x}_d(t), \vspace{2mm} \label{w1}
  \\
  \displaystyle
  &\hspace{-0mm}&w_2(t) := \ddot{x}_d(t) + \frac{b_f+b}{M} \dot{x}_d(t) - \frac{1}{M}w_f(t),
    \label{w2}
\end{eqnarray}
\end{subequations}
with $w_f$ as in \eqref{Fd}.
The perturbations $w_i(t)$, $i = \{1,2\}$, are bounded due to \cref{assumption: xd}.
All system parameters are positive, implying that in \eqref{switched system general structure}, for $i\in\{1,2\}$, $K_i,B_i>0$ and $A_i$ is Hurwitz.
The environment is located at $x=0$, so switching occurs at $x = x_d(t)-z_1 =0$.
Expressed in the $z$-coordinates, the free motion and contact subspaces, respectively denoted by $\Omega_1$ and $\Omega_2$, are time-varying: $\Omega_1(t) := \{ z \in \mathds{R}^2 | x_d(t) - z_1 \leq 0 \}$ and
$\Omega_2(t) := \{ z \in \mathds{R}^2 | x_d(t) - z_1 > 0\}$. Note that for all $t$,
$\Omega_1(t) \cup \Omega_2(t) = \mathds{R}^2$
and $\Omega_1(t) \cap \Omega_2(t) = \emptyset$.

The environment stiffness $k_e$ is typically much higher than the control gain $k_p$. Furthermore, the true value of $k_e$ and $b_e$ are usually unknown and therefore the control parameters cannot be selected to result in $K_1=K_2$ and $B_1=B_2$ in \eqref{switched system general structure}. Thus, in general,  $\Sigma^p$ in \eqref{switched system general structure} represents a switched system.
The stability of $\Sigma^p$ does not follow from the stability of each of the two continuous subsystems (corresponding to free motion and contact) taken separately, as shown, e.g., in \cite{Raibert1981,Chiaverini1993} (see also \cite{Liberzon2003} in the scope of generic switched systems).
Hence, the switching between the two subsystems, corresponding to making and breaking contact, must also be taken into account.
This is the purpose of the next section. 
\section{Stability analysis} \label{sec: stability analysis}
In this section, sufficient conditions are provided under which $\Sigma^p$ in \eqref{switched system general structure} is input-to-state stable (ISS) with respect to the input $w_i(t)$, $i = \{1,2\}$. Note that $w_i(t)$ depends $x_d(t)$, thereby encoding the information of $F_d(t)$ during the contact phases.

The following definitions, taken from \cite{Biemond2010}, are required for the stability analysis.
\begin{definition}
  Consider a region $\mathcal{T}_i \subset \mathds{R}^2$. If $z\in \mathcal{T}_i$ implies $cz\in \mathcal{T}_i$, $\forall c\in(0,\infty)$ and $\mathcal{T}_i \backslash \{0\}$ is connected, then $\mathcal{T}_i$ is a \textbf{cone}.
\end{definition}
\begin{definition} \label{definition: visible eigenvectors}
  Let $\dot{z} = A_i z$  be the dynamics on an open cone $\mathcal{T}_i \subset \mathds{R}^2$, $i = 1, ...,m$. An \textbf{eigenvector} of $A_i$ is \textbf{visible} if it lies in $\overline{\mathcal{T}}_i$, the closure of $\mathcal{T}_i$.
\end{definition}

As a stepping stone towards proving ISS of \eqref{switched system general structure}, we provide sufficient conditions for the global uniform exponential stability (GUES) of the origin of $\Sigma^p$ when $w_i \equiv 0$. This corresponds to studying the unperturbed system
\begin{equation}
    \Sigma^u: \quad
  \dot{z} = A_i z  \quad \forall z\in \Omega_i(t). \label{switched system general structure no perturbation}
\end{equation}
The GUES of the origin of $\Sigma^u$ for any $x_d(t)$ satisfying \cref{assumption: xd} can be concluded by considering the worst-case switching sequence \cite{Liberzon2003,Margaliot2006}.
In this way, we obtain the time-invariant system $\Sigma^w$, defined below, with state-based switching, that represents the worst-case switching sequence for $\Sigma^u$ in \eqref{switched system general structure no perturbation}.
The worst-case switching sequence is defined as the switching sequence that results in the slowest convergence (or fastest divergence) of the solution of $\Sigma^u$ towards (or from) the origin.
Denote with $\sigma(t): \mathds{R} \rightarrow \{ 1,2\}$ the switching sequence corresponding to  $i \in \{1,2\}$ in \eqref{switched system general structure no perturbation}.
Note that $\sigma(t)$ depends on the initial condition $z(t_0)=z_0$.
Then, the solution of $\Sigma^u$ starting from $z_0$ at $t_0$ will be denoted by $z(t) = \Phi_u(t,t_0;\sigma) z_0$, with $\Phi_u(t,t_0;\sigma)$ the state transition matrix associated with the switching sequence $\sigma(t)$.
For $K_2>K_1$, representing a manipulator interacting with a stiff environment,
the worst-case dynamical system $\Sigma^w$, associated with the worst-case switching sequence, is characterized by the following \nolinebreak[4]lemma.
\begin{lemma} \label{lemma: worst case switching}
  Consider the switched system
  \begin{equation}
    \Sigma^w: \quad
  \dot{z} = A_i z, \qquad \forall z\in \mathcal{S}_{i}, \label{worst case switched system}
  \end{equation}
  with $A_1$ and $A_2$ as in \eqref{switched system general structure}. Assume $K_2 > K_1$  and let
  \begin{align}
    \mathcal{S}_{1} &= \{ z \in \mathds{R}^2 | z_2( (K_1-K_2) z_1 + (B_1-B_2) z_2 ) \leq 0 \}, \nonumber \\
    \mathcal{S}_{2} &= \{ z \in \mathds{R}^2 | z_2( (K_1-K_2) z_1 + (B_1-B_2) z_2 ) > 0 \}. \nonumber
  \end{align}
  For the solution of $\Sigma^u$ in \eqref{switched system general structure no perturbation} corresponding to an arbitrary switching signal $\sigma(t)$ and initial condition $z_0$, $\|\Phi_u(t,t_0;\sigma)z_0\|\leq \|\Phi_{w}(t,t_0) z_0\|$ for $t \geq t_0$, where $\Phi_w$ denotes the state transition matrix of $\Sigma^w$ in \eqref{worst case switched system}. In this sense, we will refer to $\Phi_w(t,t_0)z_0$, $t \geq t_0$, as the worst-case response of $\Sigma^u$ with initial condition $z_0$.
\end{lemma}
\textbf{Proof:}
  Let $\dot z = A_{\sigma(t)} z$ denote the time-varying vector field associated with the switching signal $\sigma(t) \in \{1,2\} ~\forall t$ corresponding to an arbitrary $x_d(t)$ satisfying \cref{assumption: xd}.
  Let $V = \frac{1}{2} z^T z$ be a positive definite comparison function, with time derivative $\dot{V} = z^T \dot{z} = z^T A_{\sigma(t)} z$. Let us define $\dot{V}_i := z^T A_i z$ for $i = \{1,2\}$. Then it holds that
  $\dot{V} = z^T A_{\sigma(t)} z  \leq \mbox{max} \left( \dot{V}_1, \dot{V}_2 \right)$.
  From the structure of $A_1$ and $A_2$ in \eqref{switched system general structure}, with $K_2> K_1$, it follows that $\dot{V}_1 > \dot{V}_2$ if $z_2( (K_1-K_2) z_1 + (B_1-B_2) z_2 ) < 0$ and vice versa, such that a switching logic based on $i = \mbox{argmax}_{j\in\{1,2\}} \dot{V}_j$, is equivalent with the one in \eqref{worst case switched system}. For equal initial conditions $z_0$, it follows that $V(\Phi_u(t,t_0;\sigma)) \leq V(\Phi_{w}(t,t_0))$.
  Since $V(z) = \frac{1}{2}  \|z\|^2$, it follows that $\|\Phi_u(t,t_0;\sigma)\|\leq \|\Phi_{w}(t,t_0)\|$ and $\Sigma^w$ generates the worst-case response of $\Sigma^u$.
\hspace{\fill} \qed

\begin{figure}
    \centering
    \includegraphics[width = 7cm]{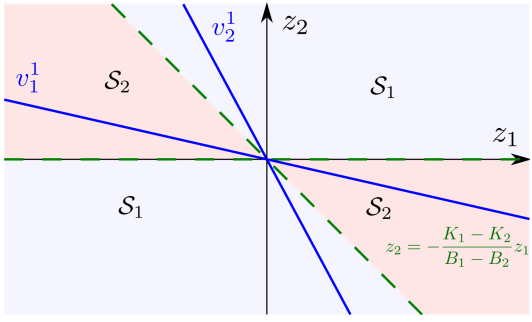}
    \caption{Switching surfaces and domains of $\Sigma^w$ for $K_2 > K_1$ and $B_2>B_1$. The vectors $v_1^1$ and $v_2^1$ represent the real eigenvectors of $A_1$ in \eqref{eigenvectors}.}
    \label{fig: worst case switching with visible eigenvector2}
\end{figure}

From the definition of $\mathcal{S}_1$ and $\mathcal{S}_2$ given in \cref{lemma: worst case switching}, we obtain the two switching surfaces $z_2 = 0$ and  $(K_1-K_2) z_1 + (B_1-B_2) z_2=0$ that characterize the worst-case switching.
These switching surfaces and the subsystems of $\Sigma^w$ that are active between the switching surfaces are visualized in \cref{fig: worst case switching with visible eigenvector2} for $K_2 > K_1$ and $B_2>B_1$.

In \cref{proposition: GUES} below, necessary and sufficient conditions for the global uniform asymptotic stability (GUAS) of $\Sigma^w$ are given. We then show in \cref{lemma: GUES}, that GUAS of $\Sigma^w$ implies GUES of $\Sigma^u$ and this, in turn, implies ISS of $\Sigma^p$ w.r.t. $w_i$ for an arbitrary $x_d(t)$ satisfying \cref{assumption: xd}. This result is given in \cref{proposition: ISS} at the end of this section and, together with \cref{proposition: GUES}, constitute the main result of this paper.
We refer the interested reader to Appendix \ref{appendix: GUAS convewise linear system} for further details about the background material used to obtain the following results.
\begin{theorem} \label{proposition: GUES}
  Let $K_i,B_i>0$, $\Delta K := K_1-K_2 <0$ and $\Delta B := B_1-B_2$.
  The origin of the unperturbed, conewise linear system $\Sigma^w$ is GUAS if at least one of the following conditions is satisfied:
  \begin{enumerate}[i.]
    \item $\Sigma^w$ has a visible eigenvector associated with an eigenvalue $\lambda<0$; in other words, one of the following two conditions is satisfied:
        \begin{enumerate}
          \item a visible eigenvector exists in $\mathcal{S}_1$, i.e., $\Delta B < 0$, $B_1^2\geq4K_1$ and $\displaystyle{\frac{\Delta K}{\Delta B} < \frac{2K_1}{B_1-\sqrt{B_1^2-4K_1}}}$
          \item a visible eigenvector exists in $\mathcal{S}_2$, i.e., $B_2^2\geq4K_2$ and one of the following conditions is satisfied:
          \begin{enumerate}[1)]
            \item $\Delta B < 0$ and $\displaystyle{\frac{\Delta K}{\Delta B} > \frac{2K_2}{B_2+\sqrt{B_2^2-4K_2}}}$, or
            \item $\Delta B \geq 0$.
          \end{enumerate}
        \end{enumerate}

        \item $\Sigma^w$ has no visible eigenvectors and $\Lambda_1\Lambda_2<1$, where $\Lambda_{i}$, $i = \{1,2\}$, are given by:
        \begin{enumerate}[1)]  
          \item if \emph{$B_i^2 < 4 K_i$},
          \begin{equation} \hspace{-5mm}
            \Lambda_i = \left( \frac{K_i}{\omega_i} \left( \frac{(\Delta K)^2}{L^2} + \frac{Q^2}{4\omega_i^2 L^2}  \right)^{-1/2} \right)^{(-1)^i} e^{-\frac{B_i}{2\omega_i} \varphi_i} \label{Lambda_i case 1}
          \end{equation}
          with $\varphi_i := \mod \left(- \arctan(\frac{ (-1)^i 2 \omega_i\Delta K}{Q}),\pi \right)$, $Q := B_i\Delta K - 2K_i\Delta B$,
          $\omega_i := \frac{1}{2}\sqrt{4K_i-B_i^2}$ and $L := \sqrt{(\Delta K)^2 + (\Delta B)^2}$.

        \item if \emph{$B_i^2 = 4 K_i$},
          \begin{equation}
          \Lambda_i = \left| \frac{B_i L}{2\Delta K -B_i\Delta B} \right|e^{\left( (-1)^{i} \frac{2\Delta K}{2\Delta K -B_i\Delta B} \right)}. \label{Lambda_i case 3}
          \end{equation}

        \item  if \emph{$B_i^2 > 4 K_i$},
          \begin{eqnarray}
            \Lambda_i &=& \left| \frac{\Delta K \lambda_{bi} + K_i \Delta B}{K_i L} \right|^{\left( (-1)^i \frac{  \lambda_{ai}}{\lambda_{bi}-\lambda_{ai}} \right)} \nonumber \\
            &&\cdot
            \left| \frac{\Delta K \lambda_{ai} + K_i \Delta B}{K_i L} \right|^{\left( (-1)^i \frac{ \lambda_{bi}}{\lambda_{ai}-\lambda_{bi}}\right)} \label{Lambda_i case 2}
          \end{eqnarray}
          with $\lambda_{ai} := \frac{-B_i-\sqrt{B_i^2-4K_i}}{2}$ and \\$\lambda_{bi} := \frac{-B_i+\sqrt{B_i^2-4K_i}}{2}$.
        \end{enumerate}
  \end{enumerate}
\end{theorem}
\textbf{Proof:}
From \cref{lemma: no unstable sliding modes} in Appendix \ref{appendix: GUAS convewise linear system} it follows that $\Sigma^w$ in \eqref{worst case switched system} has no sliding modes on the switching surfaces.
Therefore, \cref{theorem: Benjamin} can be applied to conclude GUAS of the origin of $\Sigma^w$. To this end, consider the conditions under points \emph{i} and \emph{ii} sequentially:
\begin{enumerate}
  \item[\emph{i.}] Since $K_i,B_i>0$, both $A_1$ and $A_2$ are Hurwitz, such that $\Re(\lambda^i_{1,2})<0$, with $\lambda^i_{1,2} = \frac{-B_i \pm \sqrt{B_i^2-4K_i}}{2}$ being the eigenvalues of $A_i$.
      An eigenvector is visible in $\mathcal{S}_i$ if the eigenvalues $\lambda_{1,2}^i$ of $A_i$ are real and for at least one of the corresponding eigenvectors
       \begin{equation} \hspace{-1mm}
          v_1^i \hspace{-1mm}:=\hspace{-1mm} \bma{c} \frac{-B_i+\sqrt{B_i^2-4K_i}}{2K_i} \\ 1 \ema,
          v_2^i \hspace{-1mm}:=\hspace{-1mm} \bma{c} \frac{-B_i-\sqrt{B_i^2-4K_i}}{2K_i} \\ 1 \ema \hspace{-5mm}\label{eigenvectors}
       \end{equation}
     it holds that $v_j^i \in \mathcal{S}_i$, with $j=1$ or $j=2$. These eigenvectors lie in the second and fourth quadrant of the phase portrait.
     For $j=1$, \cref{fig: worst case switching with visible eigenvector2} shows the eigenvectors $v_1^1$ and $v_2^1$ and switching surfaces $z_2 = 0$ and $z_2  = -\frac{K_1-K_{2}}{B_1-B_{2}} z_1$.
    The subsystem active in $\mathcal{S}_1$ has a visible eigenvector if $\Delta B <0$ (switching surface in second and fourth quadrant) and the slope of the corresponding real eigenvector with the steepest slope, i.e. $v_1^1$, is steeper than $z_2  = -\frac{K_1-K_{2}}{B_1-B_{2}} z_1$, i.e. the inequalities of condition \emph{i.(a)} of the theorem  hold.

    Similarly, it follows that the subsystem active in $\mathcal{S}_2$ has a visible eigenvector if either 1) $\Delta B < 0$ (switching surface in second and fourth quadrant) and $z_2  = -\frac{K_1-K_{2}}{B_1-B_{2}} z_1$ has a steeper slope than the real eigenvector of $\mathcal{S}_2$ with the least steep slope, i.e. $v_2^2$, or 2) $\Delta B \geq 0$ (switching surface in first and third quadrant, hence $\mathcal{S}_2$ spans at least the whole second and fourth quadrant). These two cases hold when conditions 1) and 2) of condition \emph{i.(b)} of the theorem are satisfied.
    For both cases, GUAS of the origin follows from case \emph{(i)} of \cref{theorem: Benjamin} in \cref{appendix: GUAS convewise linear system}. \vspace{1mm}

\item[\emph{ii.}] In case no visible eigenvectors exist, case \emph{(ii)} of \cref{theorem: Benjamin}, provided in \cref{appendix: GUAS convewise linear system}, must hold with $\Lambda := \Lambda_1^2\Lambda_2^2<1$, or equivalently, $\Lambda_1\Lambda_2<1$ in order for the origin of $\Sigma^w$ to be GUAS. The expressions \eqref{Lambda_i case 1}-\eqref{Lambda_i case 2} follow from the three cases \eqref{Lamda i case 1}-\eqref{Lamda i case 2} of part \emph{(ii)} of \cref{theorem: Benjamin}, with the following vectors and matrices
    \begin{eqnarray}
    \rho_{12}^1 = -\rho_{12}^2 = \bma{c}1 \\ 0\ema, \quad\rho_{21}^1 = \rho_{21}^2 = \frac{1}{L}\bma{c}\Delta B \\ -\Delta K\ema , \nonumber
    \end{eqnarray}
      1)  $P_i = \bma{cc} \frac{-K_i}{\omega_i} & \frac{-B_i}{2\omega_i} \\ 0 & 1\ema$,~
      2)  $P_i = \bma{cc} -\frac{2}{B_i} & -\frac{4}{B_i^2} \\ 1 & 0\ema$,\\
      3)  $P_i = \bma{cc} \frac{\lambda_{ai}}{K_i} & \frac{\lambda_{bi}}{K_i} \\ 1 & 1\ema$. \hspace{\fill} \qed
\end{enumerate}
This Theorem can be interpreted as follows.
If the system $\Sigma^w$ does not have a visible eigenvector (case \emph{ii}), the response spirals around the origin and visits the regions $\mathcal{S}_1$ and $\mathcal{S}_2$ infinitely many times.
In such a case, the worst-case system $\Sigma^w$ switches between free motion and contact, but if $\Lambda < 1$, defined in the proof of \cref{proposition: GUES}, the resulting bouncing behavior is asymptotically stable, implying that the amplitude of the oscillation decays over time.
Furthermore, since the trajectory leaves each cone in \emph{finite} time (see \cref{lemma: no visible eigenvector} in Appendix \ref{appendix: GUAS convewise linear system}), the time between two switches is fixed and finite, implying that Zeno behavior (infinitely many switches in finite time) of $\Sigma^w$ is excluded.
If $\Sigma^w$ does have a visible eigenvector with $\lambda<0$ (case \emph{i}), the response converges to the origin exponentially without leaving the cone (see \cref{lemma: visible eigenvector} in  Appendix \ref{appendix: GUAS convewise linear system}). Then, the system does not switch between free motion and contact and bouncing of the manipulator against the environment does not occur.

The following lemma states that GUAS of $\Sigma^w$ implies GUES of $\Sigma^u$.
\begin{lemma} \label{lemma: GUES}
  If $\Sigma^w$ in \eqref{worst case switched system} is GUAS, then the origin of $\Sigma^u$ in \eqref{switched system general structure no perturbation} is GUES for arbitrary $x_d(t)$ satisfying \cref{assumption: xd}.
\end{lemma}
\textbf{Proof:}
    By \cref{lemma: worst case switching},
    $\|\Phi_u(t,t_0;\sigma)z_0\|\leq \|\Phi_{w}(t,t_0)z_0\|$. So, if the origin of $\Sigma^w$ is GUAS,
    then so is the origin of $\Sigma^u$ for arbitrary $x_d(t)$ satisfying \cref{assumption: xd}. Then, from Theorem 2.4 of \cite{Liberzon2003} it follows that the origin of $\Sigma^u$ is GUES for arbitrary $x_d(t)$ satisfying \cref{assumption: xd}.
\hspace{\fill} \qed

From \cref{lemma: GUES}, $\Sigma^u$ is GUES if $\Sigma^w$ is GUAS, and this last fact is guaranteed when one of the conditions given in \cref{proposition: GUES} holds true.
The following theorem provides conditions for ISS of the perturbed system $\Sigma^p$ in \eqref{switched system general structure}.
\begin{theorem} \label{proposition: ISS}
  Consider the perturbed system $\Sigma^p$ in \eqref{switched system general structure}, with piecewise-continuous, bounded input $w_i(t)$. If the origin of the unperturbed system $\Sigma^u$ in \eqref{switched system general structure no perturbation} is GUES for arbitrary $x_d(t)$ satisfying \cref{assumption: xd},
  then $\Sigma^p$ is ISS w.r.t. $x_d(t)$.
\end{theorem}
\textbf{Proof:}
  For an arbitrary switching sequence $\sigma(t): \mathds{R} \rightarrow \{ 1,2\}$, resulting from arbitrary $x_d(t)$ satisfying \cref{assumption: xd}, the solution of $\Sigma^p$, with initial condition $z_0$ at $t_0$, can be expressed as (see \cite{Sun2005}, Chapter 1)
  \begin{equation}
    z(t) = \Phi_u(t,t_0;\sigma) z_0 + \int_{t_0}^t \Phi_u(t,\tau;\sigma) ~ N w_i(\tau) d\tau.  \label{solution perturbed linear switched system}
  \end{equation}
  If the origin of $\Sigma^u$ is GUES, which is guaranteed if the conditions in \cref{lemma: GUES} are satisfied, $\|\Phi_u(t,t_0;\sigma)\| \leq c e^{-\lambda(t-t_0)} $, for some  constants $c, \lambda>0$. Then, it follows from \eqref{solution perturbed linear switched system} that
  \begin{eqnarray}
    \|z(t)\| &\leq& \| \Phi_u(t,t_0;\sigma) z_0 \| + \| \int_{t_0}^t \Phi_u(t,\tau;\sigma) N w_i(\tau) d\tau \| \nonumber \\
    &\leq& c e^{-\lambda(t-t_0)}  \|z_0\| + c \int_{t_0}^t e^{-\lambda(t-\tau)}   ~ \|N w_i(\tau)\| d\tau \nonumber \\
    &\leq& \underbrace{c e^{-\lambda(t-t_0)}  \|z_0\|}_{\beta(\|z_0\|,t-t_0)} + \underbrace{ \frac{c}{\lambda} \sup_{t_0\leq\tau\leq t} \|N w_i(\tau)\|}_{\gamma(\sup_{t_0\leq\tau\leq t} \|N w_i(\tau)\|)}. \nonumber
  \end{eqnarray}
  Since $\beta$ is a class $\mathcal{KL}$ function and $\gamma$ is a class $\mathcal{K}$ function, $\Sigma^p$ is ISS for arbitrary $x_d(t)$ satisfying \cref{assumption: xd}.
\qed

This theorem can be interpreted as follows. If $N w_i(t) \equiv 0$, the response of $\Sigma^p$ is equivalent to the response of $\Sigma^u$, whose origin is GUES. Due to \eqref{Fd}, $x_d(t)$ encodes the information of $F_d(t)$ during the contact phase, so $x \rightarrow x_d(t)$ and $F_e \rightarrow F_d(t)$ exponentially. If $N w_i(t) \neq 0$, the response of $\Sigma^p$ deviates from the response of $\Sigma^u$, (i.e. $x$ and $F_e$ will only converge to neighbourhoods of $x_d(t)$ and $F_d(t)$, respectively), but due to the ISS property the response of $\Sigma^p$ is bounded and the bound on the error norm $\|z\|$, with $z$ defined in \eqref{states}, will depend on the norm of the perturbation $N w_i$. 
\section{Example with a stiff environment} \label{sec: stiff environment}
\begin{figure}
\centering
\includegraphics[trim = 1mm 2mm 3mm 0mm,clip,width=83mm]{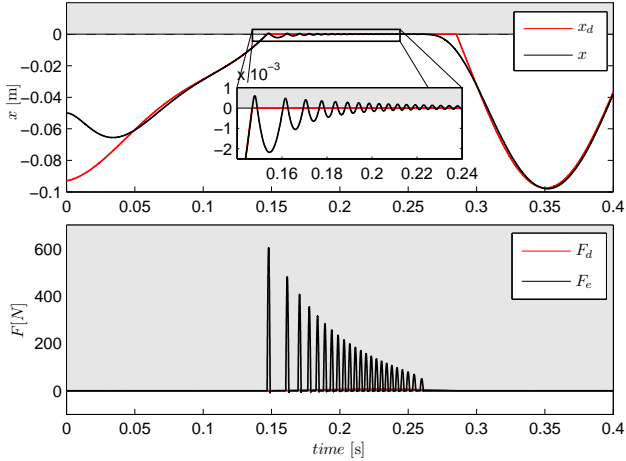}
\vspace{-2mm}
\caption{Simulation results with $b_f = 5$. The grey area indicates the contact phase. }
\label{fig: sim_bouncing}
\end{figure}

\begin{figure}
\centering
\includegraphics[trim = 1mm 2mm 3mm 0mm,clip,width=85mm]{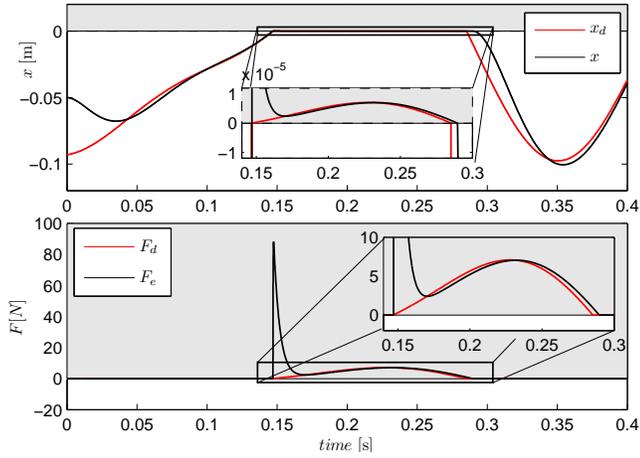}
\vspace{-2mm}
\caption{Simulation results with $b_f = 9000$. The grey area indicates the contact phase.}
\label{fig: sim_no_bouncing}
\end{figure}

We now illustrate the use of the developed theory by means of simulations and show the implications of satisfying \cref{proposition: ISS} on the controller design. Consider a manipulator with $M=1$ kg and $b = 0$ Ns/m (i.e. no viscous friction is present in the manipulator to help dissipate energy), interacting with an environment with $k_e = 10^6$ N/m and $b_e = 10$ Ns/m.
For the control parameters we choose $M_a = 0.8$ kg, $k_p = 4000$, $k_d = 80$, $k_f = 1$ and $b_f = 5$.
For this parameter set, the eigenvectors of $A_2$ in \eqref{switched system general structure} are complex, such that no visible eigenvectors exist in the contact phase (see \cref{definition: visible eigenvectors}). The eigenvectors of $A_1$ in \eqref{switched system general structure} are real, but not visible.
The response of the system is shown in \cref{fig: sim_bouncing}.
Although $x_d(t)$ and $F_d(t)$ used for the simulation in \cref{fig: sim_bouncing} are not necessarily worst-case inputs, the value $\Lambda = \Lambda_1^2 \Lambda_2^2 = 10.16$ indicates that the system is \emph{potentially} unstable (the conditions in case \emph{ii} of \cref{proposition: GUES} are \emph{necessary and sufficient} for stability of $\Sigma^w$, since they are based on its exact solution).
The controller tracks $x_d(t)$ in free motion, but due to the stiff environment and nonzero impact velocity, a large peak force occurs (see bottom plot in \cref{fig: sim_bouncing}). The manipulator bounces then back from the environment and breaks contact. During the 0.15 s of intended contact, the manipulator continues to bounce and is not able (see \cref{fig: sim_bouncing}) to track the desired contact force $F_d(t)$, which has a maximum of 7 N. Around 0.27 s the \emph{motion} controller is no longer able to bring the manipulator in contact with the environment due to the relatively large negative derivative term in \eqref{controlller free motion}. The amplitude of the bouncing does decay over time, but \cref{fig: sim_bouncing} clearly illustrates an undesired response.
The problem is the lack of damping in contact.
Increasing the damping level in the force controller to $b_f = 9000$ results in $\Lambda = \Lambda_1^2 \Lambda_2^2 = 0.98$, such that the origin of $\Sigma^w$ is GUAS (see \cref{proposition: GUES}) and the system $\Sigma^p$ is ISS, for \emph{any} motion-force profile $x_d(t), F_d(t)$ satisfying \cref{assumption: xd} (see \cref{proposition: ISS}).
With $b_f = 9000$, the manipulator does not bounce against the environment (see \cref{fig: sim_no_bouncing}) and, after the peak impact force, the contact force $F_e$ approximately tracks $F_d(t)$.

However, such a high damping gain $b_f$ in contact is probably not realizable in practice, so therefore we propose a different solution, namely a compliant manipulator. The results of \cref{proposition: ISS} are then used as a systematic procedure to design the stiffness of the wrist.
This solution is discussed in the next section.


\section{Compliant manipulator design} \label{sec: compliant manipulator}
This section discusses the motivation for the need of a compliant manipulator and shows how \cref{proposition: ISS} can be used to tune the stiffness and damping properties of the introduced compliancy.

\subsection{Motivation and design} \label{sec: compliant manipulator, subsec: motication and design}
A drawback of the high damping gain $b_f$ used in the simulation in \cref{fig: sim_no_bouncing} is that it results in a lag in tracking $F_d(t)$ for $t\in[0.17,0.28]$ (sluggish response).
Moreover, most manipulators are not equipped with velocity sensors. So typically, the velocity signal $\dot{x}$, used in \eqref{controller contact}, must be obtained from the position measurements. Due to measurement noise, encoder quantization and a finite sample interval, 
realizing the damping force $-b_f \dot x$ appearing in \eqref{controller contact} is very hard, for not saying impossible, in practice, even if one would use a state observer to estimate $\dot x$.

\begin{figure}
    \centering
    \includegraphics[width = 5.5cm]{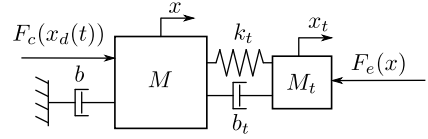}
    \caption{Manipulator with compliant wrist.}
    \label{fig: block diagram manipulator with compliant tool}
\end{figure}

Inspired by the favorable properties of the skin around a human finger,
we propose, as a more practical alternative, to design the manipulator by including passive compliance in the connection between the arm and the end-effector (wrist) as sketched in \cref{fig: block diagram manipulator with compliant tool}.
Indicating 
with $k_t$ and $b_t$, respectively, the stiffness and damping coefficient of the wrist and with $x_t$ the position of the end-effector, the dynamics of the compliant manipulator is
\begin{subequations}
\label{compliant dynamics}
\begin{align}
  M \ddot{x} + b \dot{x} &= F_c - F_t,  \label{x dyn}\\
  M_t \ddot{x}_t &= F_t - F_e(x_t,\dot{x}_t), \label{x_t dyn} 
\end{align}
\end{subequations}
where the internal force $F_t$ is given by
\begin{equation}
  F_t = k_t ( x - x_t) + b_t (\dot{x} - \dot{x}_t). \label{F_t}
\end{equation}
The environment model and controller are still given by \eqref{Kelvin-Voigt} and \eqref{Fc}, respectively, and \eqref{Fc} controls $x$ to $x_d(t)$.


The compliant wrist and end-effector are designed to improve the response during and after the impact phase. So, we consider a design where the mass $M_t$ is smaller than $M$ to reduce the kinetic energy of $M_t$ engaged at impact. The damping $b_t$ is larger than $b_e$ to help dissipate the impact energy and provide more damping in the contact phase. The stiffness $k_t$ is designed smaller than $k_e$ ($k_e$ is much larger than all other parameters) to reduce the eigenfrequency and increase the damping ratio of the contact phase. In symbols, we can write these assumptions as
\begin{equation}
  M_t \ll M,~ k_t \ll k_e, ~b_t \gg b_e, \mbox{ and } \frac{b_t}{k_e} \ll 1 \mbox{ s}. \label{conditions parameter values}
\end{equation}

\subsection{Reduced order model}
The stability results of \cref{sec: stability analysis} only apply to two-dimensional systems.
The dynamics of the 2-DOF compliant manipulator of \eqref{compliant dynamics} is 4-dimensional, so \cref{proposition: GUES} cannot be applied directly. However, when \eqref{conditions parameter values} is satisfied, the compliant 2-DOF manipulator \eqref{compliant dynamics} exhibits a clear separation between fast and slow dynamics. In free motion, the fast dynamics are related to $x - x_t$, and, in contact, to the end-effector position $x_t$.
The time-scale of the (exponentially stable) fast dynamics is very small compared to the time-scale of interest, so the slow dynamics can be considered as the dominant dynamics describing the response $x$ of the compliant manipulator to the control input $F_c(t)$.

Consider the 2-DOF compliant manipulator \eqref{compliant dynamics},  \eqref{Kelvin-Voigt} with  $M \sim 10^{0}$, $b \sim 10^{0}$, $M_t \sim 10^{-2}$, $k_t \sim 10^4$, $b_t \sim 10^2$, $k_e \sim 10^6$ and $b_e \sim 10^1$.
The model reduction analysis in \cref{appendix: proof model reduction} shows that the slow time-scale response of this system in free motion and contact considered separately can be approximated by the following model of reduced ($2^{\mbox{nd}}$) order:
\begin{equation}
  M \ddot{x} + b \dot{x} = F_c - \bar{F}_e(x,\dot x), \label{reduced compliant dynamics}
\end{equation}
\begin{equation}
  \bar{F}_e(x,\dot x) = \left\{ \begin{array}{ll}
     0 & \mbox{for } x \leq 0 \\
     \bar{b}_e \dot{x}  + \bar{k}_e x & \mbox{for } x > 0
  \end{array} \right.  \label{Kelvin-Voigt 2}
\end{equation}
with $\bar{b}_e := b_t \frac{k_e}{k_t+k_e}$ and $\bar{k}_e := k_t \frac{k_e}{k_t+k_e}$. The fraction $\frac{k_e}{k_t+k_e} \approx 1$ for $k_t \ll k_e$, so $k_t$ and $b_t$ directly influence the perceived environment damping and stiffness by the mass $M$.

The reduced-order dynamics \eqref{reduced compliant dynamics}, \eqref{Kelvin-Voigt 2} are obtained separately for the free motion and contact case. During free motion to contact transitions, the high-frequency dynamics of \eqref{compliant dynamics}, \eqref{Kelvin-Voigt}, which are not captured in \eqref{reduced compliant dynamics}, \eqref{Kelvin-Voigt 2}, might still be excited. However, the simulations provided in \cref{sec: compliant manipulator subsec: example} indicate that the response of \eqref{reduced compliant dynamics}, \eqref{Kelvin-Voigt 2} accurately approximates the response of \eqref{compliant dynamics}, \eqref{Kelvin-Voigt}, subject to \eqref{conditions parameter values} and controlled by \eqref{Fc}. Hence, we claim that the reduced-order model \eqref{reduced compliant dynamics}, \eqref{Kelvin-Voigt 2} can be used to analyze stability of \eqref{compliant dynamics}, \eqref{Kelvin-Voigt}, in closed loop with \eqref{Fc}.

\subsection{Stability of the reduced-order model}
Since the reduced-order model \eqref{reduced compliant dynamics}, \eqref{Kelvin-Voigt 2} has exactly the same structure as \eqref{dynamics manipulator}, \eqref{Kelvin-Voigt}, we can employ the stability analysis as in \cref{sec: stability analysis} to design the parameters of the controller in \eqref{Fc}.
In contact, we use a similar expression to relate $F_d(t)$ to $x_d(t)$, namely
\begin{equation}
 F_d(t) = \bar{k}_e x_d(t) + \bar{b}_e \dot{x}_d(t) + \bar{w}_f(t), \quad \mbox{for } F_d(t)>0
 \label{Fd_bar}
\end{equation}
with $\bar{w}_f(t):= (\tilde{k}_e - \bar{k}_e) x_d(t) + (\tilde{b}_e - \bar{b}_e) \dot{x}_d(t)$, and $\tilde{k}_e$ and $\tilde{b}_e$ available estimates of $\bar{k}_e$ and $\bar{b}_e$, respectively.
The design of the desired trajectories such that $x_d(t)$ is bounded  and twice differentiable is discussed in \cref{appendix: design desired trajectories}.

The system described by \eqref{reduced compliant dynamics}, \eqref{Kelvin-Voigt 2}, \eqref{Fc} and \eqref{Fd_bar} can be expressed in the form $\Sigma^p$ of \eqref{switched system general structure}, with \eqref{K1 and B1}, \eqref{w1}, \eqref{w2} and 
\begin{equation}
  K_2:= \frac{(1+k_f)\bar{k}_e}{M},  \quad
B_2:= \frac{(1+k_f) \bar{b}_e+b_f+b}{M}. \\
\end{equation}
As a result, ISS can be concluded from \cref{proposition: ISS} for arbitrary $x_d(t)$ satisfying \cref{assumption: xd} if the conditions of \cref{proposition: GUES} are satisfied. Compared to the system without compliant wrist, we now have more flexibility to tune the parameters for stability and performance. From \cref{proposition: GUES} we can compute the required values of the design parameters $k_t$ and $b_t$ to meet design specifications such as the existence of a visible eigenvector corresponding to a stable eigenvalue (implying bounceless impact) or an upper bound on $\Lambda = \Lambda_1^2 \Lambda_2^2$ in \cref{proposition: GUES}.
In case of a visible eigenvector corresponding to a stable eigenvalue, stable contact with the environment \emph{without} bouncing can be achieved for \emph{all} bounded signals $x_d(t),F_d(t)$.

\subsection{Compliant manipulator example} \label{sec: compliant manipulator subsec: example}
\begin{figure}
\centering
\includegraphics[trim = 1mm 0mm 3mm 0mm,clip,width=83mm]{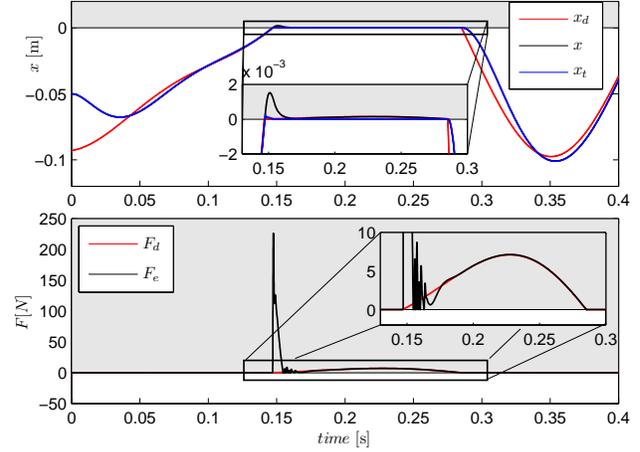}
\vspace{-2mm}
\caption{Simulation results of compliant manipulator described by \eqref{compliant dynamics}. The grey area indicates the contact phase.}
\label{fig: sim_tip}
\end{figure}

The following example illustrates how to design the compliant wrist parameters $M_t$, $b_t$ and $k_t$ to improve the closed-loop performance compared to the simulation results of \cref{fig: sim_bouncing}. For the design of the end-effector, consider $M_t = 0.05$ kg and $k_t = 5 \cdot 10^4$ N/m ($k_t \ll k_e$, but still large to minimize the spring-travel in the wrist). With $b_f = 5$ Ns/m, we require $b_t > 170$ Ns/m to guarantee that $\Lambda <1$, such that one of the conditions of \cref{proposition: GUES} is satisfied. \cref{fig: sim_tip} shows the response of the \emph{unreduced} compliant system \eqref{compliant dynamics}, \eqref{Fc} and \eqref{Kelvin-Voigt}, with $b_t = 171$ Ns/m.
Compared to \cref{fig: sim_bouncing}, the peak impact force is reduced. During the first 20 ms of intended contact, the tip makes and breaks contact due to the fast dynamics of \eqref{compliant dynamics}. After 20 ms the fast dynamics of \eqref{compliant dynamics} damp out, the slow dynamics become dominant and the response of \eqref{compliant dynamics} converges to that of \eqref{reduced compliant dynamics}. Hence, $F_e$ tracks the desired trajectory $F_d(t)$ (without a sluggish response as in \cref{fig: sim_no_bouncing}).
Since stability is now obtained with a (more practical) passive implementation, there is more freedom in tuning the parameters of the controller in \eqref{Fc}.

\begin{figure}
\centering
\includegraphics[trim = 1mm 0mm 3mm 0mm,clip,width=83mm]{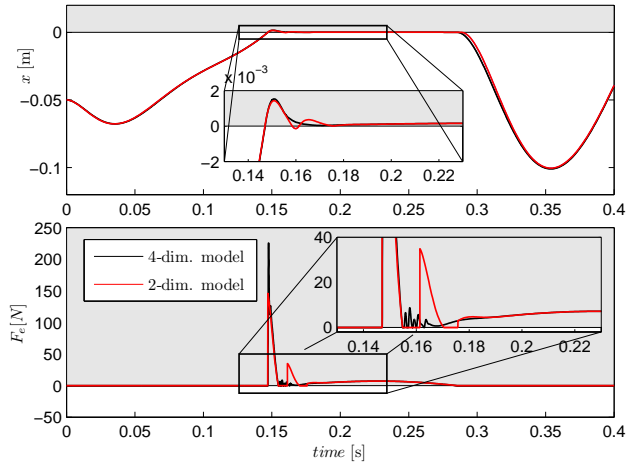}
\vspace{-2mm}
\caption{Simulation results of the compliant manipulator described by \eqref{compliant dynamics}, \eqref{F_t}, \eqref{Kelvin-Voigt} (black line), and the reduced-order model described by \eqref{reduced compliant dynamics}, \eqref{Kelvin-Voigt 2} (blue line). The grey area indicates the contact phase, i.e. $x_t>0$ for the 4-dimensional model \eqref{compliant dynamics}, \eqref{F_t}, \eqref{Kelvin-Voigt} and $x>0$ for the 2-dimensional model \eqref{reduced compliant dynamics}, \eqref{Kelvin-Voigt 2}.}
\label{fig: sim_ comparison}
\end{figure}

Finally, \cref{fig: sim_ comparison} shows a comparison of the response of the 4-dimensional compliant manipulator described by \eqref{compliant dynamics}, \eqref{F_t}, \eqref{Kelvin-Voigt}, controlled by \eqref{Fc}, and the 2-dimensional model described by \eqref{reduced compliant dynamics}, \eqref{Kelvin-Voigt 2}, and controlled by \eqref{Fc}.
The peak impact force of the 2-dimensional model is 30 percent smaller, but the time of making and breaking contact is almost equal. The main difference between the two models is found between 0.155 s and 0.18 s, where the fast dynamics of the 4-dimensional model are excited due to bouncing of the tip against the environment. Here, the 2-dimensional model has a second peak around 0.16 s due to a larger impact velocity compared to the tip of the 4-dimensional model.
After 0.18 s, the response of both models is similar, indicating that \eqref{reduced compliant dynamics}, \eqref{Kelvin-Voigt 2} is indeed a good (slow time-scale) approximation of \eqref{compliant dynamics}, \eqref{F_t}, \eqref{Kelvin-Voigt} and that \cref{proposition: GUES} can be used as a guideline for the design of damping and stiffness parameters of the compliant wrist and of the switching controller \eqref{Fc}.

\subsection{Discussion}
From the expressions $\bar{k}_e$ and $\bar{b}_e$ in \eqref{Kelvin-Voigt 2} and the results in \cref{fig: sim_tip}, we see that the compliance in the manipulator can contribute to guaranteeing stability and improve the tracking performance during free motion to contact transitions.
In fact, with $b_t\gg b_e$, the end-effector acts as a vibration-absorber, dissipating the kinetic energy present at impact.
And due to the compliance, we can lower the stiffness and increase the damping of the perceived manipulator-environment connection in contact. As a result, the controllers \eqref{controlller free motion} and \eqref{controller contact} can be tuned separately for optimal performance in free motion and contact, respectively, rather than a trade-off to guarantee stability during transitions in case of a rigid manipulator.
Using a light end-effector and tuning of $b_t$ and $k_t$ to satisfy \cref{proposition: GUES}, stable contact with the environment can be made for arbitrary $x_d(t)$ and $F_d(t)$ satisfying \cref{assumption: xd}. Moreover, if a visible eigenvector exists in the contact phase, even bouncing of the manipulator can be prevented for arbitrary $x_d(t)$ and $F_d(t)$.

\section{Conclusion} \label{sec: conclusion}
We consider the position-force control of a manipulator in contact with a stiff environment, focusing on a single direction of contact interaction.
We propose a novel switching controller that, when tuned properly, ensures stable bounded tracking of \emph{time-varying} motion and force profiles. Moreover, we provide sufficient conditions for the input-to-state stability (ISS) of the closed-loop tracking error dynamics
with respect to perturbations.  
The stability analysis that we introduce in this paper shows that for realistic parameter values, a high level of controller damping is required during contact to guarantee stability of the closed-loop system.
Such high-gain velocity feedback is
undesirable for achieving satisfying tracking performance and, moreover, likely unrealizable in practice.
Based on the results of our investigation, we propose to combine the proposed switching controller with a mechanical design of the manipulator that includes a compliant wrist.
The stability conditions presented in \cref{proposition: GUES,proposition: ISS} can be used as a guideline for the design of the damping and stiffness of this compliant wrist as well as the control parameters to guarantee stability.
Furthermore, by designing the closed-loop response to possess visible eigenvectors, those stability conditions can be used to shape the closed-loop response to prevent
persistent bouncing of the manipulator against the environment for arbitrary desired motion and force profiles.

\bibliographystyle{unsrt}        
\bibliography{bibfile}           
%
\appendix
\section*{Appendix}
\label{appendix}
\section{Design of continuous signals $x_d(t)$ and $\dot{x}_d(t)$} \label{appendix: design desired trajectories}
In this appendix, we present a method to obtain continuous signals $x_d(t)$ and $\dot{x}_d(t)$ (and corresponding $F_d(t)$), required as reference signals for the switched controller \eqref{Fc}, from the continuous and bounded reference profiles $\tilde{x}_d(t)$ and $\tilde{F}_d(t)$ \emph{specified by the user}.

\begin{figure}
\centering
\includegraphics[trim = 0mm 0mm 3mm 0mm,clip,width=50mm]{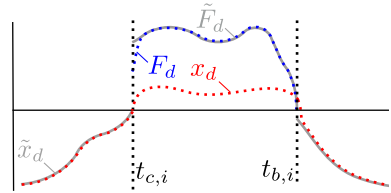}
\caption{Example of a construction of $x_d(t)$ and $F_d(t)$ from $\tilde{x}_d(t)$ and $\tilde{F}_d(t)$.}
\label{fig: construct xd}
\end{figure}

Denote the $i^{th}$ intended time of making contact by $t_{c,i}$ and the subsequent time of breaking contact by $t_{b,i}$ respectively, as indicated in \cref{fig: construct xd}.
Then, during the contact time interval $[t_{c,i}, t_{b,i}]$, $F_d(t)$ and $x_d(t)$ are obtained from
\begin{subequations}
\label{filter xd Fd in contact}
\begin{align}
  \ddot{y}_1 &= - 2\gamma_1 \dot{y}_1 - \gamma_1^2 (y_1-\tilde{F}_d(t)), \label{filter Fd in contact} \\
  &  \hspace{20mm} y_1(t_{c,i})= \hat{k}_e x_d(t_{c,i}) + \hat{b}_e \dot{x}_d(t_{c,i}), \nonumber \\
  & \hspace{20mm} \dot{y}_1(t_{c,i})= \hat{k}_e \dot{x}_d(t_{c,i}) + \hat{b}_e \ddot{x}_d(t_{c,i}), \nonumber \\
  \ddot{y}_2 &= -\frac{\hat{k}_e}{\hat{b}_e} \dot{y}_2 + \frac{1}{\hat{b}_e} \dot{y}_1 , \qquad y_2(t_{c,i})= x_d(t_{c,i}),  \label{filter xd in contact} \\
  &\hspace{35mm}  \dot{y}_2(t_{c,i})=  \dot{x}_d(t_{c,i}) \nonumber
\end{align}
\end{subequations}
with the outputs $F_d(t) = y_1$, $x_d(t) = y_2$ and $\dot{x}_d(t) = \dot{y}_2$ for $t \in[t_{c,i}, t_{b,i}]$.
The $y_2$-dynamics follow from the time derivative of \eqref{Fd hat} and guarantee continuity of $x_d(t)$ and $\dot{x}_d(t)$ at $t = t_{c,i}$.
The $y_1$-dynamics represent a critically damped second-order filter on $\tilde{F}_d(t)$ to guarantee continuity of $F_d(t)$ and $\dot{F}_d(t)$ at $t = t_{c,i}$. As a  guideline, the time constant $\gamma_1>0$ in \eqref{filter Fd in contact} is chosen such that the 'bandwidth' of this filter is significantly higher than the frequencies present in $\tilde{F}_d(t)$.

Continuity of the profiles $x_d(t)$ and $\dot{x}_d(t)$ when breaking contact is guaranteed when these profiles during the free motion time interval $[t_{b,i},t_{c,i+1}]$ are obtained from $\tilde{x}_d(t)$ filtered by the critically damped second-order filter
\begin{eqnarray}
  \ddot{y}_3 &=& - 2\gamma_2 \dot{y}_3 - \gamma_2^2 (y_3-\tilde{x}_d(t)), \hspace{1mm} y_3(t_{b,i})=x_d(t_{b,i}), \hspace{3mm} \label{filter xd in free motion}\\
  && \hspace{39.5mm} \dot{y}_3(t_{b,i})= \dot{x}_d(t_{b,i}), \nonumber
\end{eqnarray}
with outputs $x_d(t) = y_3$ and $\dot{x}_d(t) = \dot{y}_3$ for $t \in[t_{b,i}, t_{c,i+1}]$. As for $\gamma_1$ in \eqref{filter Fd in contact}, the time constant $\gamma_2>0$ in \eqref{filter xd in free motion} is chosen such that the 'bandwidth' of \eqref{filter xd in free motion} is significantly higher than the frequencies typically present in $\tilde{x}_d(t)$.

\section{GUAS of a conewise linear system} \label{appendix: GUAS convewise linear system}
The stability results presented here are based on the results presented in \cite{Biemond2010} and ultimately lead to the statement of \cref{theorem: Benjamin}, which is used in the proof of \cref{proposition: GUES} in the main text of this paper.
The results in \cite{Biemond2010} apply to \emph{continuous}, conewise linear systems. The conewise linear system $\Sigma^w$ in \eqref{worst case switched system} is, however, discontinuous. The continuity of the vector field is required in \cite{Biemond2010} to exclude the existence of unstable sliding modes at the switching surfaces of the conewise linear system.
The following lemma shows that $\Sigma^w$ has no sliding modes at the switching surfaces.
\begin{lemma} \label{lemma: no unstable sliding modes}
  For $K_1-K_2 < 0$, the conewise linear system $\Sigma^w$ has no sliding mode.
\end{lemma}
\textbf{Proof:}
  The existence of a sliding mode at the two switching surfaces $z_2 = 0$ and $z_2 = -\frac{K_1-K_2}{B_1-B_2} z_1$ of $\Sigma^w$ are considered sequentially:
  \begin{itemize}
    \item Consider the subspace $\{ z\in \mathds{R}^2 | z_1 \geq 0\}$. The normal $\mathcal{N}_1$ to the switching surface $z_2 = 0$ is given by $\mathcal{N}_1 = [0,1]^T$. The inner product of the vector fields $A_i z$, $i \in \{1,2\}$, with $\mathcal{N}_1$ at the switching surface $z_2=0$ reads $\lambda \mathcal{N}_1^T A_i \nu_1 = - \lambda K_i$, where $\nu_1 = [1,0]^T$ and $\lambda \geq 0$. This inner product has the same sign for both vector fields associated with $i=1$ and $i=2$, such that no sliding mode exists at the switching surface $z_2=0$, see e.g. \cite{Leine2004}.
    \item Consider the subspace $\{ z\in \mathds{R}^2 | z_1 \geq 0\}$. The normal $\mathcal{N}_2$ to the switching surface $z_2 = -\frac{K_1-K_2}{B_1-B_2} z_1$ is given by $\mathcal{N}_2 = \frac{1}{L}[\Delta K,\Delta B]^T$, with $\Delta K:= K_1-K_2$, $\Delta B := B_1-B_2$ and $L:= \sqrt{(\Delta K)^2 + (\Delta B)^2}$.
        The projection of the vector fields $A_i z$, $i \in \{1,2\}$, with $\mathcal{N}_2$ at the switching surface $z_2 = -\frac{K_1-K_2}{B_1-B_2} z_1$ read
        \small\begin{eqnarray*}
          \lambda \mathcal{N}_2^T A_1 \nu_2 &=& \frac{\lambda}{L^2} ((\Delta K)^2 +K_1(\Delta B)^2 - B_1(\Delta K)(\Delta B)), \\
          \lambda \mathcal{N}_2^T A_2 \nu_2 &=& \frac{\lambda}{L^2} ((\Delta K)^2 +K_2(\Delta B)^2 - B_2(\Delta K)(\Delta B)),
        \end{eqnarray*} \normalsize
        \noindent where $\nu_2 = \frac{1}{L}[-\Delta B,\Delta K]^T$ and $\lambda \geq 0$.
        It can be shown that $\lambda \mathcal{N}_2^T A_1 \nu_2 - \lambda \mathcal{N}_2^T A_2 \nu_2=0$, $\forall K_i, B_i >0$, hence, the inner products $\lambda \mathcal{N}_2^T A_1 \nu_2$ and $\lambda \mathcal{N}_2^T A_2 \nu_2$ have the same sign, such that no sliding mode exists on the switching surface $z_2 = -\frac{K_1-K_2}{B_1-B_2} z_1$, see e.g. \cite{Leine2004}.
  \end{itemize}
  With a similar analysis, the same results can be obtained for the subspace $\{ z\in \mathds{R}^2 | z_1 \leq 0\}$.
\hspace{\fill} \qed

\begin{figure}
    \centering
    \includegraphics[width = 4.5cm]{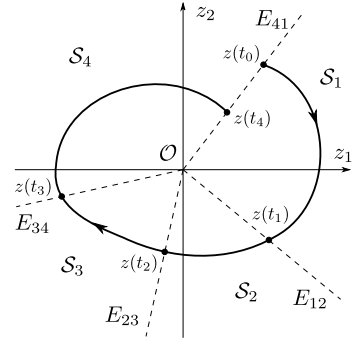}
    \caption{Example of a trajectory of \eqref{worst case switched system} that traverses each cone once per rotation.}
    \label{fig: State-space_with_cones}
\end{figure}

The following lemma holds for continuous conewise linear systems $\Sigma^w$ \emph{with}  visible eigenvectors.
\begin{lemma}[\cite{Biemond2010}]
\label{lemma: visible eigenvector}
  Consider a continuous, conewise linear system of the form $\Sigma^w$. When this system contains one or more visible eigenvectors, then $z=0$ is an asymptotically stable equilibrium of $\Sigma^w$ if and only if all visible eigenvectors correspond to eigenvalues $\lambda <0$.
\end{lemma}
This lemma can also be shown to be valid for discontinuous conewise systems $\Sigma^w$ in the absence of a sliding mode.
The following lemma is useful in the analysis of the behavior of $\Sigma^w$ in the absence of visible eigenvectors.
\begin{lemma}[\cite{Biemond2010}]
  \label{lemma: no visible eigenvector}
  Let $\bar{\mathcal{S}}_i$ be a closed cone in $\mathds{R}^2$. Suppose no eigenvectors of $A_i\in \mathds{R}^{2\times2}$ are visible in $\bar{\mathcal{S}}_i$. Then for any initial condition $z_0 \in \bar{\mathcal{S}}_i$, with $z_0 \neq 0$, there exists a time $t \geq 0$ such that $e^{A_it}z_0 \not\in \bar{\mathcal{S}}_i$.
\end{lemma}
If \cref{lemma: no visible eigenvector} holds for all cones, the trajectories exhibit a spiralling response, visiting each region $i$ once per rotation, as indicated in \cref{fig: State-space_with_cones}.
Stability for a spiraling motion can be analyzed by the computation of a return map.
Suppose the trajectory of \eqref{worst case switched system} enters a region $\mathcal{S}_i$ at $t_{i-1}$ at position $z(t_{i-1})$, which is located on the boundary $E_{i-1,i}$ between cones $\mathcal{S}_{i-1}$ and $\mathcal{S}_i$, such that $z(t_{i-1})$ can be expressed as $z(t_{i-1}) = p^i \rho_{i-1,i}$. Here, $p^i$ represents the radial distance from the origin at time $t_{i-1}$  and $\rho_{i-1,i}$ is  the unit vector parallel to the boundary $E_{i-1,i}$. The trajectory crosses the next boundary $E_{i,i+1}$ at finite time $t_i$ (\cref{lemma: no visible eigenvector}), and the position of this crossing is given by $z(t_i) = p^{i+1} \rho_{i,i+1}$, such that $z(t_i)$ is parallel to $\rho_{i,i+1}$. Since the dynamics in each cone are linear, the time $t_i$ can be computed explicitly. The crossing positions are linear in $p^i$, so expressions for a scalar $\Lambda_i$, such that $p^{i+1} = \Lambda_i p^{i}$, can be obtained.

In order to construct the return map, consider for each cone the following coordinate transformation
\begin{equation}
  \tilde{z}^i = P_i^{-1} z, \hspace{0mm} \mbox{for } \tilde{z}^i \in \tilde{\mathcal{S}}_i := \{ \tilde{z}^i \in\mathds{R}^2 ~|~ \tilde{z}^i = P_i^{-1} z ~|~ z \in \bar{\mathcal{S}}_i \}, \nonumber
\end{equation}
where $P_i$ is given by the real Jordan decomposition of $A_i$, yielding $A_i = P_i J_i P_i^{-1}$.
Depending on the eigenvalues of $A_i$, three different cases can be distinguished.
\begin{enumerate} 
  \item $A_i$ has complex eigenvalues denoted by $a_i \pm \omega_i$, where $a_i$ and $\omega_i$ are real constants and $\omega_i>0$. Then, $J_i = \big[\begin{smallmatrix}  a_i & -\omega_i \\ \omega_i & a_i \end{smallmatrix} \big]$. Define $\phi(r_1,r_2)$ to be the angle in counter clockwise direction from vector $r_1$ to vector $r_2$. Then,
      \begin{equation}
        \Lambda_i = \frac{\| \tilde{\rho}_{i-1,i}^i \|}{\| \tilde{\rho}_{i,i+1}^i \|} e^{\frac{a_i}{\omega_i} \phi(\tilde{\rho}_{i-1,i}^i~,~\tilde{\rho}_{i,i+1}^i)}, \label{Lamda i case 1}
      \end{equation}
      with $\tilde{\rho}_{i-1,i}^i := P_i^{-1}\rho_{i-1,i}$ and $\tilde{\rho}_{i,i+1}^i := P_i^{-1}\rho_{i,i+1}$
  \item $A_i$ has two equal real eigenvalues $\lambda_{ai}$ with geometric multiplicity 1. Then, $J_i = \big[\begin{smallmatrix} \lambda_{ai} & 1 \\ 0 & \lambda_{ai} \end{smallmatrix} \big]$ and
      \begin{equation}
        \Lambda_i = \left| \frac{e_2^T \tilde{\rho}_{i-1,i}^i}{e_2^T \tilde{\rho}_{i,i+1}^i} \right| e^{\lambda_{ai} \left( \frac{e_1^T \tilde{\rho}_{i,i+1}^i}{e_2^T \tilde{\rho}_{i,i+1}^i} - \frac{e_1^T \tilde{\rho}_{i-1,i}^i}{e_2^T \tilde{\rho}_{i-1,i}^i}\right)}, \label{Lamda i case 3}
       \end{equation}
       where $e_1 := [1,0]^T$ and $e_2 := [0,1]^T$.

  \item $A_i$ has two distinct real eigenvalues $\lambda_{ai}$ and $\lambda_{bi}$. Then, $J_i = \big[\begin{smallmatrix} \lambda_{ai} & 0 \\ 0 & \lambda_{bi} \end{smallmatrix} \big]$ and
      \begin{equation}
        \Lambda_i = \left| \frac{e_2^T  \tilde{\rho}_{i,i+1}^i}{e_2^T  \tilde{\rho}_{i-1,i}^i} \right|^{\frac{\lambda{a_i}}{\lambda{b_i}-\lambda{a_i}}}
        \left| \frac{e_1^T \tilde{\rho}_{i,i+1}^i}{e_1^T \tilde{\rho}_{i-1,i}^i} \right|^{\frac{\lambda{b_i}}{\lambda{a_i}-\lambda{b_i}}}. \label{Lamda i case 2}
      \end{equation}

\end{enumerate}
From the scalars $\Lambda_i$ for each cone $\mathcal{S}_i$, $i = 1,\ldots,m$, the return map between the positions $z_k$ and $z_{k+1}$ of two consecutive crossings of the trajectory $z(t)$ with the boundary $E_{m1}$ can be computed as
$z_{k+1} = \Lambda z_k$, where
\begin{equation}
  \Lambda = \prod_{i=1}^m \Lambda_i. \nonumber
\end{equation}

\cref{theorem: Benjamin} below is an extension of Theorem 6 in \cite{Biemond2010} and provides necessary and sufficient conditions for GUAS of the origin of the discontinuous, conewise linear system $\Sigma^w$.
\begin{theorem} \label{theorem: Benjamin}
Under the assumption that no sliding modes exist, the origin of the discontinuous, conewise linear system $\Sigma^w$ in \eqref{worst case switched system} is GUAS if at least one of the following conditions is satisfied:
\begin{enumerate}
  \item[(i)] In each cone $\mathcal{S}_i$, $i = 1,...,m$, all visible eigenvectors are associated with eigenvalues $\lambda < 0$.
  \item[(ii)] In case there exists no visible eigenvector, it holds that $\Lambda<1$.
\end{enumerate}
\end{theorem}
\textbf{Proof:}
If no sliding modes exist on the switching surfaces, GUAS of the origin of the discontinuous system $\Sigma^w$ can be proven similarly to the proof of Theorem 6 in \cite{Biemond2010} for continuous, conewise linear systems.
\hspace{\fill} \qed

From \cref{lemma: no unstable sliding modes} it follows that $\Sigma^w$ in \eqref{worst case switched system} has no sliding modes on the switching surfaces, so \cref{theorem: Benjamin} can indeed be applied to conclude GUAS of the origin of $\Sigma^w$. 
\section{Model reduction compliant manipulator} \label{appendix: proof model reduction} The model \eqref{reduced compliant dynamics}-\eqref{Kelvin-Voigt 2} describes the slow dynamics of \eqref{compliant dynamics}, \eqref{F_t}, \eqref{Kelvin-Voigt} and is obtained by employing Theorem 11.2 of \cite{Khalil2002}. With this theorem, the slow dynamics are obtained for an infinite time horizon $t \in [t_0,\infty]$. We will refer to it as \emph{Tikhonov's extended theorem}, since the original theorem of Tikhonov, see e.g. Chapter 7 of \cite{Thikonov1985}, only applies on a finite time horizon $t\in[t_0,t_f]$.

Tikhonov's extended theorem is applicable to systems described by (non)linear continuous, possibly time varying, dynamics. The dynamics of \eqref{compliant dynamics}, \eqref{F_t}, \eqref{Kelvin-Voigt} are not continuous due to the switch between free motion and contact. Therefore, we consider the model reduction of the free motion ($x_t \leq 0$) and contact ($x_t > 0$) phases separately. The simulation results presented in \cref{sec: compliant manipulator} indicate that for the considered parameter values the response of the original compliant manipulator dynamics \eqref{compliant dynamics}, \eqref{F_t}, \eqref{Kelvin-Voigt}, \emph{including } the transitions between free motion and contact, can be approximated by the dynamics of the reduced-order model \eqref{reduced compliant dynamics}-\eqref{Kelvin-Voigt 2}.

Below, for both free motion and contact, the reduction of the 4$^{\mbox{th}}$-order model \eqref{compliant dynamics}, \eqref{F_t}, \eqref{Kelvin-Voigt} to the second-order model \eqref{reduced compliant dynamics}-\eqref{Kelvin-Voigt 2} is performed in two steps, where in each step the model is reduced with one order.
\vspace{5mm}

\emph{Free motion}: Consider the following states
\begin{align}
  e&:= x - x_t \nonumber\\
  \dot{e}&:= \dot{x} - \dot{x}_t. \nonumber
\end{align}
The following parameters are used as an example to illustrate the separation of the two distinct time-scales of the system described by \eqref{compliant dynamics}, \eqref{F_t}, \eqref{Kelvin-Voigt}: $M \sim 10^{0}$, $b \sim 10^{0}$, $M_t \sim 10^{-2}$, $k_t \sim 10^4$, $b_t \sim 10^2$, $k_e \sim 10^6$ and $b_e \sim 10^1$.
For these parameter values, the dynamics \eqref{compliant dynamics}, \eqref{F_t}, \eqref{Kelvin-Voigt} in free motion can be written as
\begin{subequations}
  \small
  \begin{eqnarray}
     &\ddot{x} = \underbrace{\frac{1}{M}}_{ \sim 10^{0}} F_c(t) - \underbrace{\frac{b}{M}}_{ \sim 10^{0}} \dot{x} - \underbrace{\frac{k_t}{M}}_{ \sim 10^{4}} e - \underbrace{\frac{b_t}{M}}_{ \sim 10^2} \dot{e} 
     \nonumber \hspace{35mm}  \\
     &\underbrace{\frac{M_t}{(1+M_t/M)k_t}}_{ \sim 10^{-6}} \ddot{e} = \underbrace{\frac{M_t}{M(1+M_t/M)k_t}}_{ \sim 10^{-6}} F_c(t) -  e - \underbrace{\frac{b_t}{k_t}}_{ \sim 10^{-2}} \dot{e} \nonumber  \hspace{35mm} \\ & -~ \underbrace{\frac{M_t b}{M(1+M_t/M)k_t}}_{ \sim 10^{-6}} \dot{x} .  \nonumber 
  \end{eqnarray}
\end{subequations}
Define $\mu_1 := \frac{M_t}{(1+M_t/M)k_t} \approx \frac{M_t}{M(1+M_t/M)k_t} \approx \frac{M_t b}{M(1+M_t/M)k_t}$ and $\mu_2:= \frac{b_t}{k_t}$. With these parameters, it follows that $\mu_1 \ll \mu_2$, and we obtain the following dynamics
 \begin{subequations}
  \label{dynamics not in contact}
  \begin{align}
     \ddot{x} &= \frac{1}{M} ( F_c(t) - b \dot{x} - k_t e - b_t \dot{e} ) \label{x not in contact} \\
     \mu_1 \ddot{e} &= \mu_1 F_c(t) - \mu_1 \dot{x} -  e - \mu_2 \dot{e} . \label{x_t not in contact}
  \end{align}
\end{subequations}
Note that $x$ does not appear directly in the right-hand side of \eqref{dynamics not in contact}.
Therefore, the dynamics of \eqref{dynamics not in contact} are described by the three states $(\dot{x},e,\dot{e})$ only.

In the analysis that follows, we consider $\mu_1$ and $\mu_2$ as \emph{singular perturbations} and use Tikhonov's extended theorem twice (once for $\mu_1$ and once for $\mu_2$) to obtain a model of reduced order.

Before proceeding, we first decouple the free response of \eqref{dynamics not in contact} from the forced response (due to $F_c(t)$). To this end, consider the coordinate transformation $\dot{\tilde{x}}:= \dot{x} - \dot{\bar{x}}_{F_c}(t)$, where $\bar{x}_{F_c}(t)$ is defined as the forced response of the slow dynamics of \eqref{dynamics not in contact} (i.e. for $\mu_1=\mu_2=0$) to the continuous and bounded input $F_c(t)$, such that
\begin{equation}
M \ddot{\bar{x}}_{F_c}(t) + b \dot{\bar{x}}_{F_c}(t) = F_c(t). \label{Fc(t) for free motion}
\end{equation}
Note that $\dot{\bar{x}}_{F_c}(t)$ and $\ddot{\bar{x}}_{F_c}(t)$ are continuous and bounded since $F_c(t)$ is continuous and bounded. By employing \eqref{Fc(t) for free motion}, the unforced dynamics of \eqref{dynamics not in contact} can be expressed as
\begin{subequations}
  \label{dynamics not in contact 2}
  \begin{align}
     \ddot{\tilde{x}} 
     &= \frac{1}{M} ( - b \dot{\tilde{x}} - k_t e - b_t \dot{e} )  \\
     \mu_1 \ddot{e} &= \mu_1 F_c(t) - \mu_1 (\dot{\tilde{x}} + \dot{\bar{x}}_{F_c}(t))-  e - \mu_2 \dot{e} .
  \end{align}
\end{subequations}
Since $\mu_1$ is much smaller than all other parameters, we treat it as the vanishing perturbation parameter and use Tikhonov's extended theorem to obtain a model of reduced order that describes the slow dynamics of this system. Consider $y = [y_1,y_2]^T:=[\dot{\tilde{x}},e]^T$ as the states of the slow dynamics $f_1(y,\zeta)$ and $\zeta:=\dot{e}$ as the state of the fast dynamics $g_1(t,y,\zeta,\mu_1)$ of \eqref{dynamics not in contact 2} according to
\begin{subequations}
\label{singular perturbation model 1}
\begin{align} \hspace{-2.5mm}
  \bma{c} \dot{y}_1 \\ \dot{y}_2 \ema &= \bma{c} \frac{1}{M} ( - b y_1 - k_t y_2 - b_t \zeta )\\ \zeta  \ema ~ =: f_1(y,\zeta) \\
  \mu_1 \dot{\zeta} &= \mu_1 F_c(t) - \mu_1 (y_1 + \dot{\bar{x}}_{F_c}(t))-  y_2 - \mu_2 \zeta  \nonumber\\
  &=: g_1(t,y,\zeta,\mu_1).
\end{align}
\label{full system 1}
\end{subequations}
For $\mu_1 = 0$, $\zeta = h_1(y) := -\frac{1}{\mu_2} y_2$ is the solution of  $0 = g_1(t,y,\zeta,0)$ for $y \in D_y= \mathds{R}^2$ and $v_1:=\zeta-h_1(y) \in D_{v1} =\mathds{R}$.
Let us analyze the three conditions of Tikhonov's extended theorem sequentially:
\begin{enumerate}
\item[\textbf{C1}.] 
    The functions $f_1$, $g_1$, their first partial derivatives with respect to $(y,\zeta,\mu_1)$, and the first partial derivative of $g_1$ with respect to $t$ are continuous and bounded on any compact subset $D_y\times D_{v1}$, since $F_c(t)$ is continuous and bounded. Furthermore, $h_1(y)$ and $[\partial g_1(t,y,\zeta,0)/\partial \zeta]$ have bounded first partial derivatives and $[\partial f_1(y,h_1(t,y),0)/\partial y]$ is Lipschitz in $y$.

\item[\textbf{C2}.] The slow dynamics of \eqref{full system 1}
    \begin{align} \hspace{-6mm}
    \dot{y} = f_1(y,h_1(y)) &= \bma{c} \frac{1}{M} ( - b y_1 - k_t y_2 + \frac{b_t}{\mu_2} y_2 )\\ - \frac{1}{\mu_2} y_2 \ema \nonumber \\
    &= \bma{cc} -\frac{b}{M} & \frac{b_t}{\mu_2 M}-\frac{k_t}{M} \\ 0 & -\frac{1}{\mu_2}  \ema \bma{c} y_1 \\ y_2 \ema \label{reduced system 1}
    \end{align}
    have a \emph{globally} exponentially stable equilibrium point $y=0$, since $-\frac{b}{M}$ and $-\frac{1}{\mu_2}$, representing the eigenvalues of the system matrix of the linear dynamics in \eqref{reduced system 1}, are both negative.

\item[\textbf{C3}.] With $\mu_1 \frac{dv_1}{dt} = \frac{d v_1}{d\tau_1}$ (i.e. $\tau_1:= \frac{1}{\mu_1} t$) the (linear) boundary-layer system
    \begin{align} \hspace{-4mm}
      \frac{\partial v_1}{\partial \tau_1} = g_1(t,y,v_1+h_1(y),0) &= - y_2 - \mu_2(v_1 - \frac{1}{\mu_2} y_2) \nonumber \\
      &= - \mu_2 v_1 \label{boundary-layer system 1}
    \end{align}
    has a \emph{globally} exponentially stable equilibrium point at the origin (since $\mu_2>0$), uniformly in $(t,y)$ with region of attraction $\mathcal{R}_{v1} = D_{v1} = \mathds{R}$.
\end{enumerate}
  From the conditions above, Tikhonov's extended theorem allows us to conclude that for all $t_0 \geq 0$, initial conditions $y_0 \in D_y$, $\zeta_0 \in D_{\zeta}:=\mathds{R}$,
  and sufficiently small $0 < \mu_1 < \mu_1^{*}$, the singular perturbation problem of \eqref{singular perturbation model 1} has a unique solution $y(t,\mu_1)$, $\zeta(t,\mu_1)$ on $[t_0,\infty)$, and
  \begin{align}
    y(t,\mu_1) - \bar{y}(t) = \mathcal{O}(\mu_1) \nonumber \\
    \zeta(t,\mu_1) - h_1(\bar{y}(t)) - \hat{v}_1(t/\mu_1) = \mathcal{O}(\mu_1) \nonumber
  \end{align}
  holds uniformly for $t \in [t_0,\infty)$, with initial time $t_0$, where $\bar{y}(t)$ and $\hat{v}_1(\tau)$ are the solutions of \eqref{reduced system 1} and \eqref{boundary-layer system 1}, with $\bar{y}(t_0) = y(t_0)$ and $\hat{v}_1(t_0) = \zeta(t_0) + \frac{1}{\mu_2} y_2(t_0)$ respectively. Moreover, given any $t_b > t_0$, there is $\mu_1^{**} \leq \mu_1^{*}$ such that
  \begin{equation}
    \zeta_1(t,\mu_1) - h_1(\bar{y}(t)) = \mathcal{O}(\mu_1) \nonumber
  \end{equation}
  holds uniformly for $t \in [t_b,\infty)$ whenever $\mu_1 < \mu_1^{**}$.
  Hence, on the domain $t \in [t_b,\infty)$, \eqref{full system 1} can be approximated by \eqref{reduced system 1}. Rewriting the reduced-order model \eqref{reduced system 1} as the time-invariant system
  \begin{subequations}
  \label{full system 2}
    \begin{eqnarray}
    \dot{y}_1 =  \frac{1}{M} ( - b y_1 - k_t y_2 + \frac{b_t}{\mu_2} y_2 ) ~ &:=& f_2(y_1,y_2,\mu_2)\\
    \mu_2 \dot{y}_2 = - y_2 \hspace{2mm} &:=& g_2(y_2,\mu_2), \label{g2() fast dynamics 2}
    \end{eqnarray}
  \end{subequations}
  it becomes clear that $\mu_2 = \frac{b_t}{k_t} \sim 10^{-2}$ is much smaller than all other parameters in \eqref{full system 2}. Hence, we can apply Tikhonov's extended theorem once more with $\mu_2$ considered as the singular perturbation parameter, $y_1$ the slow dynamics and $y_2$ the fast dynamics.

  The details regarding the reduction step with $\mu_2$ considered as the singular perturbation is performed in a similar fashion as the first reduction step and is therefore omitted here for the sake of brevity.  
  With $y_2 = h_2(y_1) := 0$ the solution of $0 = g_2(y_2,0)$, the following globally exponentially stable slow dynamics of \eqref{full system 2} are obtained
  \begin{equation}
    \dot{y}_1 = f_2(y_1,h_2(y_1),0) =  -\frac{b}{M} y_1.  \label{reduced system 2}
  \end{equation}
  With $v_2 := y_2-h_2(y_1)$ and $\tau_2:=\frac{1}{\mu_2} t$, the boundary-layer system $\frac{\partial v_2}{\partial \tau_2} = g_2(y_1,v_2+h_2(y_1),0) = -v_2$ is globally exponentially stable. Hence, the three conditions of Tikhonov's extended theorem are satisfied, such that it can be concluded that \eqref{reduced system 2} is an approximation of \eqref{full system 2}.
  After reversing the coordinate transformation, i.e. $y_1 = \dot{x} - \dot{\bar{x}}_{F_c}(t)$, and using \eqref{Fc(t) for free motion}, we obtain
  \begin{equation}
       M \ddot{x} + b \dot{x} = F_c(t) \label{final reduced model free motion}
  \end{equation}
  as the approximation of  \eqref{compliant dynamics} in free motion. \vspace{5mm}

\emph{Contact}: Similar as for the free motion case, the model reduction for the contact case is performed in two steps. Due to the relatively high environmental contact stiffness $k_e$ in \eqref{compliant dynamics}, it is expected that $x_t$ (and time derivatives) is approximately equal to zero (the nominal position of the environment).
Therefore, the motion $x_t$ of the tip can be considered as the fast dynamics, and the motion $x$ of the manipulator can be considered as the slow dynamics.
The dynamics \eqref{compliant dynamics}, \eqref{F_t}, \eqref{Kelvin-Voigt} for $x_t>0$ can be rewritten as
\begin{subequations}
\label{dyn contact}
\begin{align}
  \ddot{x} = \underbrace{\frac{1}{M}}_{\sim 10^0} F_c(t) - \underbrace{\frac{(b+b_t)}{M}}_{\sim 10^2} \dot{x} + \underbrace{\frac{b_t}{M}}_{\sim 10^2} \dot{x}_t - \underbrace{\frac{k_t}{M}}_{\sim 10^4} (x-x_t) \label{x dyn contact} \\
  \underbrace{\mu_3}_{\sim 10^{-8}} \ddot{x}_t = \underbrace{\frac{k_t}{k_t+k_e}}_{\sim 10^{-2}} x + \underbrace{\frac{b_t}{k_t+k_e}}_{\sim 10^{-4}} \dot{x} -  x_t - \underbrace{\frac{b_t + b_e}{k_t+k_e}}_{\sim 10^{-4}} \dot{x}_t, \label{xt dyn contact}
\end{align}
\end{subequations}
where $\mu_3 := \frac{M_t}{k_t+k_e}$.
Consider the coordinate transformation
\begin{align}
  y_1 &:= x - \bar{x}_{F_c}(t), \nonumber \\
  y_2 &:= \dot{x} - \dot{\bar{x}}_{F_c}(t), \hspace{12mm} \zeta_1 := \dot{x}_t - \dot{\bar{x}}_{t,F_c}(t), \label{coordinate transformation 3}\\
  y_3 &:= x_t - \bar{x}_{t,F_c}(t), \nonumber
\end{align}
such that $y = [ y_1,y_2,y_3]^T =0$ and $\zeta_1=0$ is the equilibrium of \eqref{dyn contact} in the new coordinates. In \eqref{coordinate transformation 3}, $\bar{x}_{t,F_c}(t)$  and $\bar{x}_{F_c}(t)$ are defined as the forced response of \eqref{dyn contact} for $\mu_3=0$,  to the continuous and bounded input $F_c(t)$, i.e.
\begin{equation}
  \dot{\bar{x}}_{t,F_c}(t) = \frac{1}{b_t+b_e} \Big( k_t \bar{x}_{F_c}(t) + b_t \dot{\bar{x}}_{F_c}(t) - (k_t+k_e) \bar{x}_{t,F_c}(t) \Big), \label{xt_bar_dot expression}
\end{equation}
\vspace{-6mm}
\begin{eqnarray}
     M\ddot{\bar{x}}_{F_c}(t) + (b+b_t)\dot{\bar{x}}_{F_c}(t) - b_t \dot{\bar{x}}_{t,F_c}(t) \hspace{15mm}\nonumber \\
     \hspace{25mm}{+}~k_t (\bar{x}_{F_c}(t) -\bar{x}_{t,F_c}(t)) = F_c(t). \label{Fc(t) for contact}
\end{eqnarray}
Using the coordinate transformation \eqref{coordinate transformation 3} and the expressions \eqref{xt_bar_dot expression} and \eqref{Fc(t) for contact},  \eqref{dyn contact} can be rewritten as
\begin{subequations}
\label{dynamics contact}
\begin{align}
  \bma{c} \dot{y}_1\\ \dot{y}_2 \\ \dot{y}_3\ema 
   &=\bma{l} y_2 \\ \frac{1}{M} ( - (b+b_t)y_2 + b_t \zeta_1 - k_t(y_1 - y_3) ) \\
   \zeta_1\ema   \nonumber \\ &=: f_3(y,\zeta_1)  \hspace{-25mm}
   \end{align}
   \begin{align}
  \mu_3 \dot{\zeta}_1 &= \frac{1}{k_t+k_e} \Big(k_t y_1 + b_t y_2  - (k_t+k_e)y_3  \nonumber \\
  &  \hspace{26mm} {-} (b_t+b_e) \zeta_1 \Big) -  \mu_3 \ddot{\bar{x}}_{t,F_c}(t) \nonumber \\
  & =: g_3(t,y,\zeta_1,\mu_3).
\end{align}
\end{subequations}
Since $\mu_3$ is much smaller than all other parameters, see \eqref{xt dyn contact},  we treat it as the vanishing perturbation and use Tikhonov's extended theorem to obtain  a model of reduced order.

The details regarding the reduction step with singular perturbation parameter $\mu_3$  follows similar to the reduction step for the free motion case with $\mu_1$ considered as the singular perturbation parameter and is therefore omitted for the sake of brevity. With
\begin{equation}
    \zeta_1 = \frac{1}{b_t+b_e} \left(k_t y_1 + b_t y_2  - (k_t+k_e)y_3 \right) =: h_3(y) \nonumber
\end{equation}
the solution of $0 = g_3(t,y,\zeta_1,0)$, the following globally exponentially stable slow dynamics of \eqref{dynamics contact} are obtained
\begin{eqnarray}
    \dot{y}  &=& f_3(y,h_3(y)) \nonumber \\
    &=& \bma{l}
     \hspace{5mm} y_2 \vspace{1mm} \\  \hdashline  \vspace{-2mm}\\ \displaystyle
     \frac{1}{M} ( - (b+b_t)y_2 + \frac{b_t}{b_t+b_e} \left(k_t y_1 + b_t y_2 \right. \\ \hspace{15mm} \left.-~ (k_t+k_e) y_3\right) -k_t y_1 + k_t y_3)
     \vspace{1mm} \\  \hdashline  \vspace{-2mm}\\  \displaystyle
     \hspace{5mm}  \frac{1}{b_t+b_e} \left(k_t y_1 + b_t y_2 -(k_t+k_e) y_3 \right)\ema
\label{reduced system contact 1}
\end{eqnarray}
With $v_3 := \zeta_1-h_3(y) \in D_{v3} = \mathds{R}$ and $\mu_3 \frac{dv_3}{dt} = \frac{d v_3}{d\tau_3}$ (i.e. $\tau_3:=\frac{1}{\mu_3} t$), the boundary-layer system
\begin{equation}
  \frac{\partial v_3}{\partial \tau_3} = g_3(t,y,v_3+h_3(y),0) = - \frac{b_t+b_e}{k_t+k_e} v_3 \nonumber
\end{equation}
is globally exponentially stable, and the conditions of Tikhonov's extended theorem are satisfied, such that it can be concluded that \eqref{reduced system contact 1} is an approximation of \eqref{dynamics contact}.
Using \eqref{xt_bar_dot expression}, \eqref{Fc(t) for contact} and inverting the coordinate transformation \eqref{coordinate transformation 3}, the (intermediate) slow dynamics \eqref{reduced system contact 1} can be written in the original coordinates as
\begin{subequations}
\label{reduced system contact 2}
\begin{eqnarray}
  M \ddot{x} 
  &=& F_c(t) - (b+b_t)\dot{x} - k_t x + k_t x_t \nonumber \\
  && {+}~\frac{b_t}{b_t+b_e} \left(k_t x + b_t \dot{x} - (k_t+k_e) x_t \right) \\
  (b_t+b_e) \dot{x}_t 
  &=& k_t x + b_t \dot{x} - (k_t+k_e) x_t. \label{aa}
\end{eqnarray}
\end{subequations}
This third-order system is further approximated to a system of order 2 by considering $\mu_4 := \frac{b_t+b_e}{k_t+k_e}$ as a singular perturbation parameter. To this end, consider the state transformation
\begin{align}
  y_1 &:= x - \bar{r}(t), \nonumber \\
  y_2 &:= \dot{x} - \dot{\bar{r}}(t), \label{coordinate transformation 4} \\
  \zeta_2 &:= (k_t+k_e)(x_t - \bar{r}_t(t)), \nonumber
\end{align}
such that $y = [y_1,y_2]^T = 0$ and $\zeta_2 = 0$ is the equilibrium of \eqref{reduced system contact 2} in the new coordinates. Here, $\bar{r}_t(t)$ is defined as the forced response of the fast dynamics of \eqref{reduced system contact 2} for $\mu_4=0$, and $\bar{r}(t)$ is defined as the forced response of the slow dynamics of \eqref{reduced system contact 2} to the input $F_c(t)$, with $\mu_4=0$, i.e.
\begin{eqnarray}
    \bar{r}_{t}(t) = k_t \bar{r}(t) + b_t \dot{\bar{r}}(t) \hspace{30mm}\label{rt_bar_dot expression} \\
    M \ddot{\bar{r}}(t) + (b+b_t \frac{k_e}{k_t+k_e})\dot{\bar{r}}(t)  + k_t \frac{k_e}{k_t+k_e} \bar{r}(t) = F_c(t). \nonumber \\
     \label{Fc(t) for contact 2}
\end{eqnarray}
Rewriting \eqref{reduced system contact 2} in terms of the coordinates $y_1$, $y_2$ and $\zeta_2$, given in \eqref{coordinate transformation 4}, and using \eqref{rt_bar_dot expression}, \eqref{Fc(t) for contact 2}, we obtain
\begin{subequations}
\label{dynamics contact 2}
\begin{align}
  \bma{c} \dot{y}_1 \vspace{5mm}\\ \dot{y}_2 \vspace{5mm}\ema 
  %
  &=\bma{l} \hspace{5mm}y_2 \\  \hdashline  \vspace{-2mm}\\
  \frac{1}{M} \Big( -(b+b_t)y_2 - k_t y_1 + \frac{k_t}{k_t+k_e} \zeta_2 \\ \hspace{15mm} {+}~\frac{b_t}{b_t+b_e} \big(k_t y_1 + b_t y_2  -  \zeta_2  \big)\Big )\ema \nonumber \\
  &=: f_4(y,\zeta_2) \\
  \underbrace{\mu_4}_{\sim 10^{-4}} \dot{\zeta}_2 
  &= \underbrace{k_t}_{\sim 10^{4}} y_1 + \underbrace{b_t}_{\sim 10^{2}} y_2 - \zeta_2 - \underbrace{\mu_4}_{\sim 10^{-4}} \dot{\bar{r}}_t(t) \nonumber \\
  &:= g_4(t,y,\zeta_2,\mu_4). \label{g4}
\end{align}
\end{subequations}
Since $\mu_4$ is small compared to the other parameters, it is considered a singular perturbation parameter for the system \eqref{dynamics contact 2} and Tikhonov's extended theorem is used once more to obtain a model of reduced order.
Again, the proof of the reduction step with $\mu_4$ considered as the singular perturbation follows similar to the previous reduction steps and is therefore omitted.

For $\mu_4 = 0$, $\zeta_2 = k_t y_1 + b_t y_2 :=h_4(y)$ is the root of $0 = g_4(t,y,\zeta_2,0)$ and the following slow dynamics of \eqref{dynamics contact 2} are obtained
\begin{eqnarray}
    \dot{y} &=& f_4(y,h_4(y)) \nonumber \\
            &=& \bma{l} \hspace{5mm} y_2 \\
        \displaystyle \frac{1}{M} \Big( - \big(b+b_t\frac{k_e}{k_t+k_e}\big)y_2  - k_t \frac{k_e}{k_t+k_e} y_1  \Big) \label{reduced system contact 3}.
      \ema
\end{eqnarray}
With $v_4 := \zeta_2-h_4(y) \in D_{v4} = \mathds{R}$ and $\mu_4 \frac{dv_4}{dt} = \frac{d v_4}{d\tau_4}$ (i.e. $\tau_4 := \frac{1}{\mu_4} t$), the boundary-layer system $\frac{\partial v_4}{\partial \tau_4} = g_4(t,y,v_4+h_4(y),0) = - v_4 $ is globally exponentially stable, and the conditions of Tikhonov's extended theorem are satisfied, such that the theorem allows us to conclude  that \eqref{reduced system contact 3} is an approximation of \eqref{dynamics contact 2}.
Using the inverse of the coordinate transformation \eqref{coordinate transformation 4}, we obtain
\small{
  \begin{equation}
    M (\ddot{x} - \ddot{\bar{r}}(t) ) = -k_t \frac{k_e}{k_t+k_e} (x - \bar{r}(t) )  - (b+b_t \frac{k_e}{k_t+k_e}) (\dot{x} - \dot{\bar{r}}(t) ). \nonumber
  \end{equation} }
  \normalsize
  Using \eqref{Fc(t) for contact 2}, the slow dynamics of \eqref{reduced system contact 2} (and thus of \eqref{dyn contact}) are given by
  \begin{equation}
   M \ddot{x} + b \dot{x} = F_c(t) - k_t \frac{k_e}{k_t+k_e} x - b_t \frac{k_e}{k_t+k_e} \dot{x}. \label{final reduced model contact} \vspace{5mm}
  \end{equation}

Finally, by combining the results \eqref{final reduced model free motion} and \eqref{final reduced model contact}, for free motion and contact, we obtain the model of reduced order described by \eqref{reduced compliant dynamics}-\eqref{Kelvin-Voigt 2}.

\end{document}